\documentclass{article} 
\usepackage{iclr2020_conference,times}

\usepackage{amsmath}
\usepackage{hyperref}
\usepackage{url}
\usepackage{comment}
\usepackage{csquotes}
\usepackage[ruled,vlined]{algorithm2e}
\usepackage{graphicx}
\graphicspath{{figures/}}
\usepackage{makecell}
\usepackage{booktabs}
\usepackage{multirow}

\title{Progressive growing of self-organized \\ hierarchical representations for exploration}

\author{Mayalen Etcheverry \& Pierre-Yves Oudeyer \& Chris Reinke \\
Flowers Team, Inria, France\\
\texttt{\{mayalen.etcheverry,pierre-yves.oudeyer,chris.reinke\}@inria.fr}
}

\iclrfinalcopy 
\begin{document}

\maketitle

\begin{abstract}
Designing agent that can autonomously discover and learn a diversity of structures and skills in unknown changing environments is key for lifelong machine learning. A central challenge is how to learn incrementally representations in order to progressively build a map of the discovered structures and re-use it to further explore. To address this challenge, we identify and target several key functionalities. First, we aim to build lasting representations and avoid \textit{catastrophic forgetting} throughout the exploration process. Secondly we aim to learn a diversity of representations allowing to discover a \enquote{diversity of diversity} of structures (and associated skills) in complex high-dimensional environments. Thirdly, we target representations that can structure the agent discoveries in a coarse-to-fine manner. Finally, we target the reuse of such representations to drive exploration toward an \enquote{interesting} type of diversity, for instance leveraging human guidance. Current approaches in state representation learning rely generally on monolithic architectures which do not enable all these functionalities. Therefore, we present a novel technique to progressively construct a \textit{Hierarchy of Observation Latent Models} for \textit{Exploration Stratification}, called \textit{HOLMES}. This technique couples the use of a dynamic modular model architecture for representation learning with intrinsically-motivated goal exploration processes (IMGEPs). The paper shows results in the domain of automated discovery of diverse self-organized patterns, considering as testbed the experimental framework from \cite{reinke2019intrinsically}.
\end{abstract}

\section{Introduction}
\label{sec_1}

Maintaining, fine-tuning and expanding the acquired knowledge of a learning agent in a continual way  is a central challenge in reinforcement learning. Despite success of recent work in reinforcement learning to master complex tasks, current artificial agents still lack the necessary autonomy and versatility to properly interact with realistic environments \citep{santucci2019intrinsically}.

Exploration, or the ability of a learning agent to autonomously discover and reach a diversity of possible states in an unknown environment, is a key ingredient for lifelong machine learning. Inspired from developmental mechanisms observed in humans, \enquote{intrinsically-motivated} or \enquote{curiosity-driven} exploration \citep{oudeyer2007intrinsic,baldassarre2013intrinsically} proposes to endow the learning agent with motivational signals to guide the search toward novel states, skills or goals. Such intrinsic rewards aim to generate a curriculum for \enquote{intelligently} exploring the environment and accumulate a repertoire of diverse (re-)usable skills \citep{forestier2017intrinsically}. Coupled with (goal-conditioned) reinforcement learning policies, intrinsically-motivated algorithms have enabled agents to acquire autonomously diverse skill repertoires that can be re-used to solve efficiently downstream tasks \citep{pathak2017curiosity,mohamed2015variational,eysenbach2018diversity}, and to maintain diverse competences in non-stationary environments \citep{colas2018curious}. While several works have studied these approaches with agents that perceive their environment at the pixel-level \citep{bellemare2016unifying} and self-generate their own goals \citep{nair2018visual,pong2019skew,reinke2019intrinsically}, their efficiency relies on the ability to learn low-dimensional state/goal spaces that can adequately represent the different factors of variations of the environment. One key challenge is how to learn representations that will enable efficient exploration in environments where these underlying factors are initially unknown and may change as the agent discovers new areas, new objects, or new ways to interact with the environment. 

Representation learning, and more specifically unsupervised feature learning, aims to automatically recover the underlying low-dimensional explanatory factors of complex observations data \citep{bengio2013representation}. Replacing the need for human hand-designed features, they are particularly suited to encode high-dimensional observations into compact latent code and hence define a goal space $\mathcal{G}$. Deep generative models have the additional advantage to generate a distribution of \enquote{plausible} latent points from which new unseen goals can easily be sampled. Recent work in goal-directed exploration extensively reuses different variants of such models as variational auto-encoders (VAEs) \citep{pere2018unsupervised,ha2018world,ha2018recurrent,caselles2018continual,caselles2019s,nair2018visual,nair2019contextual,reinke2019intrinsically}, generative adversarial networks (GANs) \citep{florensa2017automatic, kurutach2018learning}, noise-contrastive estimation of mutual information \citep{anand2019unsupervised} and autoregressive methods \citep{ostrovski2017count}. The representation $R$ is either pretrained before exploration \citep{pere2018unsupervised}, or learned incrementally \citep{nair2018visual,pong2019skew,reinke2019intrinsically}, or from a generative replay model \citep{caselles2018continual,caselles2019s}. However, they all rely on a monolithic representation model $R$ to recover all the factors of variations, preventing the agent to actively organize the discoveries in different modules and at different levels of granularity. Even though the use of a replay can mitigate the phenomenon of catastrophic forgetting, such architecture generally lacks flexibility to encode new \textit{types} of information, i.e. to learn diverse representations associated to diverse kinds of structures, and to adapt to the environment increasing complexity.

In this paper, we propose a novel method to give the agent more versatility to augment and structure its world model representation and reuse it for the goal sampling strategy. Following the intuition of \cite{elman1993learning} on the importance of \enquote{starting small} both on the task data distribution and on the network memory capacity, we propose to actively grow a hierarchy of embedding networks (deep generative models such as VAEs) as the agent is discovering novel structures in its environment. The agent starts with a small network capacity and can incrementally augment it by freezing an existing module and splitting it into two child modules with their own capacity, preventing by construction the phenomenon of \textit{catastrophic forgetting}. The tree-structured representation unsupervisedly partitions the observations into distinct branches leading to a hierarchy of specialized goal space representations. Moreover, by encoding observations (and hence goals) at different levels of granularity, the proposed architecture automatically produces an \textit{exploration stratification} that can target discovery of a \enquote{diversity of diversity}. As a proof of concept, we use as test-bed environment a continuous game of life where diverse visual structures can self-organize. We compare the discoveries of IMGEPs equipped with different goal space representations: a fixed-architecture VAE and the proposed adaptive architecture HOLMES. We also implemented as use-case of our architecture an algorithm that leverages the learned structure to guide exploration toward a \textit{desired type} of diversity. \\
Our contributions are twofold. First, we introduce a dynamic modular model architecture for representing the \enquote{diversity of diversity} present in complex environments. This is, to our knowledge, the first work that proposes to progressively grow the capacity of the agent visual world model into an organized hierarchical representations. Secondly, we propose to leverage the structure of the hierarchy to guide the exploration toward a certain \textit{type} of diversity, opening interesting perspectives for the integration of a human evaluator in the loop.
\begin{figure}[b!]
\begin{center}
\includegraphics[width=\linewidth]{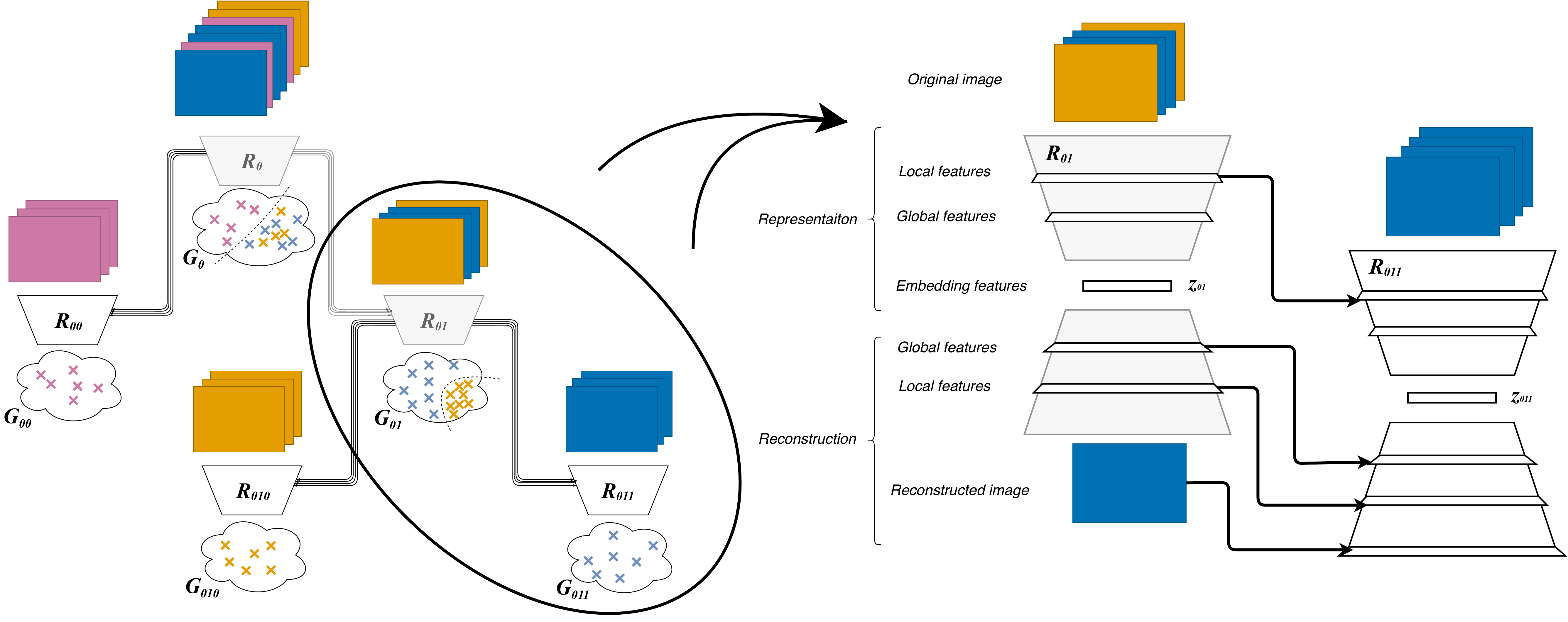}
\end{center}
\vspace{-5pt}
\caption{Hierarchy of Observation Latent Models for Exploration Stratification (HOLMES).}
\label{fig_1}
\end{figure}
\section{Approach}
\label{sec_2}
We first explain the architectural approach for learning representations provided that the agent receives an input flow of observations. Then we explain how it can be coupled with intrinsically-motivated goal exploration processes (IMGEP) where the data is collected by the exploring agent.

\paragraph{Hierarchical Observation Latent Models}
The model architecture takes inspiration from \textit{Progressive Neural Networks} (PNN) \citep{rusu2016progressive}, a dynamic model architecture that was proposed for continual learning and applied to a given sequence of reinforcement learning tasks. PNN explicitly prevent \textit{catastrophic forgetting} by instantiating a new neural network (column) for each new task, and support \textit{transfer} between tasks by connecting the new column to all the previously trained columns via learned \textit{lateral connections}. In the following, we explain the different modifications made to adapt PNN to deep generative models in the context of continual state representation learning, which remains an unexplored area \citep{lesort2019generative}. \\
The global \textit{sequential} architecture is modified into a \textit{hierarchical} representational architecture $\mathcal{R}$. The hierarchy starts with a single root neural network $\mathcal{R}_0$. When a \textit{saturation} signal is triggered, the parameters of $\mathcal{R}_0$ are frozen and two child networks $\mathcal{R}_{00}$ and $\mathcal{R}_{01}$ are instantiated. Input observations $x$ forward first through $\mathcal{R}_0$ and are then send to one child network based on a \textit{boundary} criterion defined in the feature space of $\mathcal{R}_0$. Each time a node gets \textit{saturated}, the split procedure is repeated in that node, resulting in a progressively deeper hierarchy of specialized goal spaces. \\
We replace the \enquote{column} network with a VAE composed of an encoder and a decoder network. To mitigate the growing number of parameters, \enquote{lateral connections} are only created between a node and its ancestors and between a reduced number of layers (original, local, global, and embedding levels). The connection scheme is summarized in figure \ref{fig_1}. Transfer is beneficial in the decoder network so a child module can reconstruct \enquote{as well as} its parent, however connections are removed between encoders as new complementary type of features should be learned. We preserve connections only at the local feature level, as CNN first layers tend to learn similar features \citep{yosinski2014transferable}. Connections between convolutional layers are defined as convolutions with $1 \times 1$ kernel. \\
Finally, at the difference of \cite{rusu2016progressive}, the extension into deeper levels of refinement is automatically handled during exploration, removing the need for a predefined sequence of tasks. 

\paragraph{Exploration Stratification}
IMGEPs are goal-oriented exploration processes, operating in a given goal space $\mathcal{G}$ which is computed by a an encoding function $\mathcal{R}$. We combine HOLMES with IMGEPs by replacing $\mathcal{R}$ with the proposed modular hierarchy $\{\mathcal{R}_k\}$. Henceforth, IMGEP operates in a hierarchy of goal spaces $\{\mathcal{G}_k\}$ and the agent has an additional degree of control in the goal sampling strategy by selecting first a goal space to explore and then a goal in that space. In this paper we considered two setups for the goal space sampling strategy: 1) the target goal space $\mathcal{G}_k$ is sampled uniformly over the tree leafs 2) After each split in the hierarchy, we \enquote{pause} exploration and assign a fixed probability to each leaf goal space. This second variant is intended to simulate the integration of a human evaluator in the loop that could, by visually browsing the current results made by the agent (see appendix \ref{sm:sec_A} for a possible visualisation) assign a \textit{score} to each goal space. During exploration, the agent selects one of the leaf goal spaces with softmax sampling on the assigned probabilities. Then, we follow \cite{reinke2019intrinsically} for sampling a goal $g$ in the selected space. \\
The other way round, the IMGEP influences the training of HOLMES by generating the data distribution and splits in the hierarchy.  For evolving HOLMES, we trigger a \textit{saturation} signal when the population of a goal space go past a threshold of $N_{max}$ points and use the reconstruction performance to create a \textit{boundary} $B_k$ in the goal space (see appendix \ref{sm:sec_B_2}). For details on IMGEP and the integration of HOLMES we refer to appendix \ref{sm:sec_B}.

\section{Experimental Results}
\label{sec_3}
We use the same experimental testbed as \cite{reinke2019intrinsically}. The environment is a continuous Game of Life, Lenia \citep{chan2018lenia}, where a variety of visual structures can self-organize but still are difficult to discover by manual parameter tuning, making it an interesting testbed for pattern exploration algorithms. We compare \textbf{IMGEP-VAE} equipped with a monolithic high-capacity VAE and \textbf{IMGEP-HOLMES} equipped with the proposed hierarchy of smaller-capacity VAEs and where the goal space selection is done uniformly over the leaf nodes. Additionally, using the classifiers from \cite{reinke2019intrinsically} to categorize the patterns of Lenia as \enquote{animals} or \enquote{non-animals}, we implemented two \textit{guided} variants where we assume that an external evaluator is interested in discovering a diversity of animals (or non-animals) patterns. Each time a split is triggered, the leaf nodes of the hierarchy get scored with the number of animals (or non-animals) patterns they currently contain, serving as basis for the softmax goal space sampling strategy. The variants are denoted \textbf{IMGEP-HOLMES(A)} (guided toward animals) and \textbf{IMGEP-HOLMES(NA)} (guided toward non- animals).

\paragraph{Can HOLMES represent a \enquote{diversity of diversity}?}
We use \textit{Representational Similarity Analysis} (RSA), technique coming from systems neuroscience \citep{kriegeskorte2008representational}, to compare the different goal spaces representations. Given the set of encoder representations learned with the IMGEP-VAE and IMGEP-HOLMES variants, and an independent set of lenia patterns (750 images), we compute the RSA matrix between all pairs of encoders. We refer to appendix \ref{sm:fig_9} for computation details and the full matrix result. Figure \ref{fig_1} shows the dissimilarity between the goal space learned by IMGEP-VAE versus the modular goal spaces learned by IMGEP-HOLMES. The results indicate a high similarity between the representations learned by the VAE and the root node HOLMES 0 (which can be seen as an \textit{early} frozen version of the VAE). This suggests that, although the VAE is additionally trained on new unseen patterns, the monolithic representation does not significantly update the \textit{type} of encoded information/diversity. However, the RSA matrix shows strong dissimilarities between HOLMES different nodes representations (see figure \ref{sm:fig_9} in appendix), confirming our intuition that HOLMES can better encode a \enquote{diversity of diversity} by learning different sets of features per node. 

\begin{figure}[h!]
\begin{center}
\includegraphics[width=\linewidth]{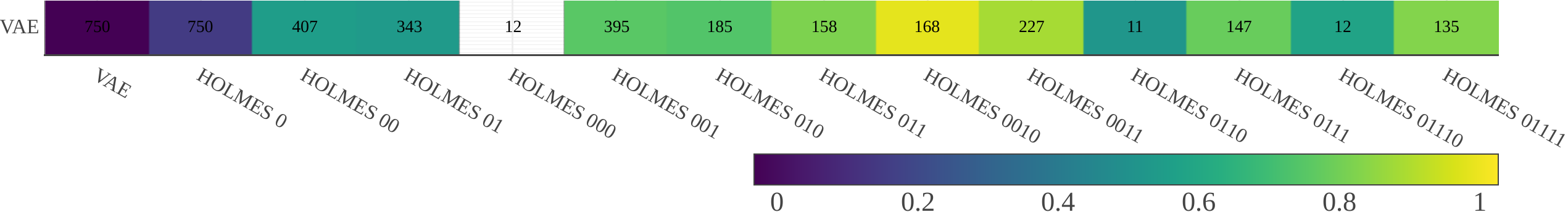}
\end{center}
\vspace{-10pt}
\caption{RSA heatmap showing disagreement (colorscale) among the different goal space representations. Numbers represent the lenia patterns (over 750) shared between each pair of goal spaces.}\label{fig_2}
\end{figure}

\paragraph{Can HOLMES drive exploration toward an \enquote{interesting type} of diversity?} Table \ref{table_1} reports the percentage of identified patterns by the different IMGEP-HOLMES variants. The results show that the IMGEP-HOLMES(A) variant (resp IMGEP-HOLMES(NA)) is finding more animals (resp non-animals) patterns, confirming that HOLMES modular architecture can be exploited to drive exploration toward a desired type of diversity. We refer to appendix \ref{sm:sec_A_1} for a qualitative illustration of these results and appendix \ref{sm:sec_A_2} for additional statistical analysis on the diversity.

\begin{table}[b!]
\caption{Comparison of percentage, across three categories, of discovered patterns for each IMGEP-HOLMES variant. For each algorithm 5 repetitions of the exploration experiment were conducted.}
\label{table_1}
\begin{center}
\vspace{-2pt}
\begin{tabular}{llll}
 & animal patterns & non-animal patterns & dead patterns
\\ \hline 
\bf IMGEP-HOLMES &  15.4 $\pm$ 2.4 & 62.2 $\pm$ 2.3 & 22.4 $\pm$ 0.8 \\
\bf IMGEP-HOLMES(A)  & \textbf{26.5 $\pm$ 3.8} & 45.9 $\pm$ 3.7 & 27.7 $\pm$ 1.1\\
\bf IMGEP-HOLMES(NA)  & 4.9 $\pm$ 0.4  & \textbf{79.6 $\pm$ 3.4} & 15.5 $\pm$ 3.1\\
\end{tabular}
\end{center}
\end{table}

\section{Conclusion}
\label{sec_4}
We presented a hierarchical model architecture for incremental learning of goal space representations, with core functionalities for dealing with open-ended environments. Specifically, it prevents the phenomenon of catastrophic forgetting, can be adaptively augmented to encode new type of information, and self-organize the agent discoveries in hierarchically organized modules. Moreover, by combining the representational architecture with intrinsically-motivated goal exploration, we showed that our approach can target discovery of a \enquote{diversity of diversity} and that the exploring agent can exploit the learned structure to efficiently drive exploration. This work opens interesting perspectives to leverage human guidance for exploration in complex systems. Future direction of research should analyze further the capabilities and limits of this architecture and consider experiments that directly integrate a human end-user.

\clearpage
\bibliography{iclr2020_conference}
\bibliographystyle{iclr2020_conference}

\appendix

\section{Additional Results}
\label{sm:sec_A}
This appendix complements the results presented in section \ref{sec_3} of the main paper. It provides visualizations of the discovered patterns in IMGEP-HOLMES (section \ref{sm:sec_A_1_1}), IMGEP-HOLMES(A) (section \ref{sm:sec_A_1_2}) and IMGEP-HOLMES(NA) (section \ref{sm:sec_A_1_3}). In addition, section \ref{sm:sec_A_2} complements the statistical results presented in the main paper.

\subsection{Qualitative Results}
\label{sm:sec_A_1}
\begin{figure}[h!]
\begin{center}
\includegraphics[angle=0,width=\linewidth]{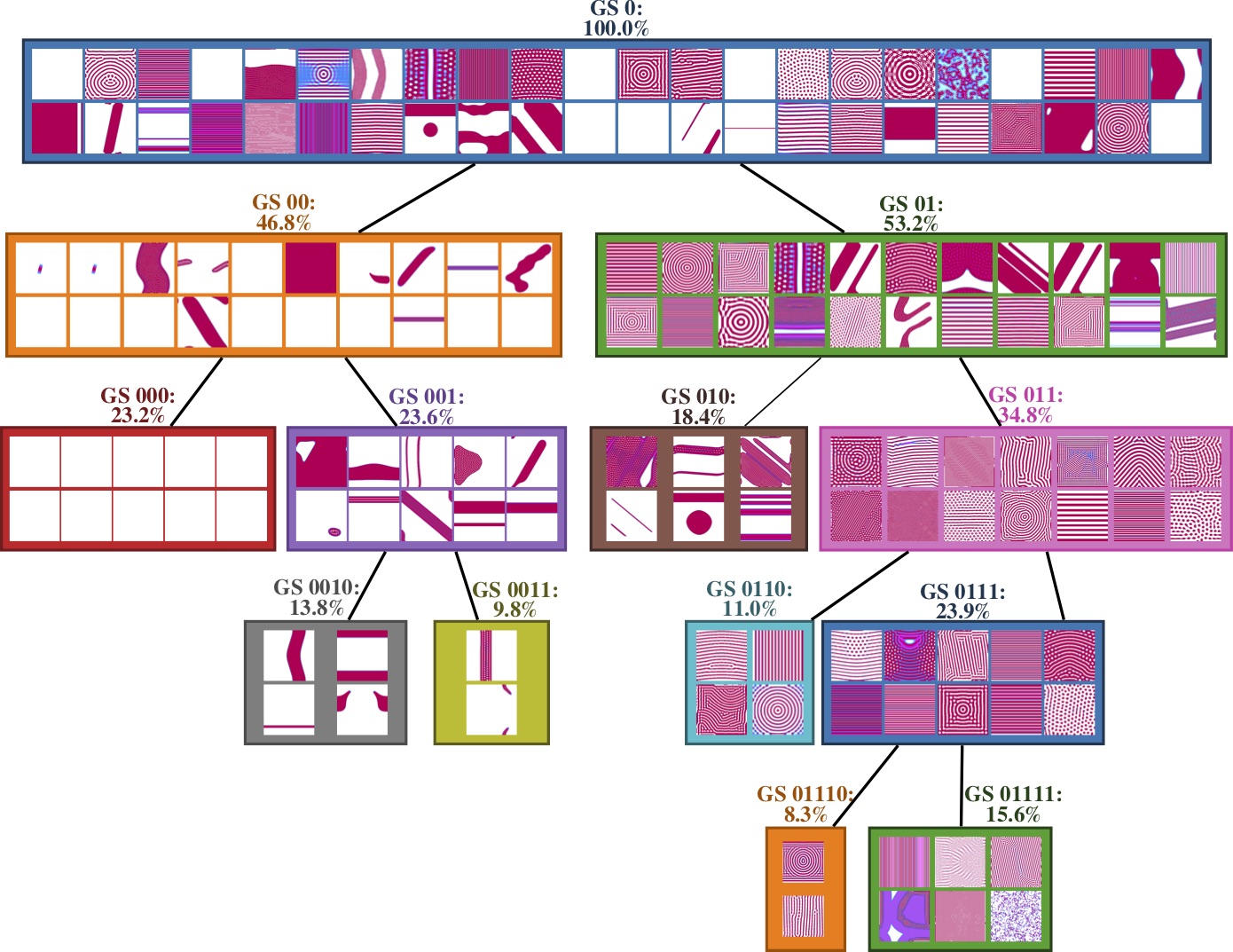}
\end{center}
\caption{Tree constructed by the IMGEP-HOLMES algorithm during a single exploration with 5000 iterations. We display (randomy selected) discovered pattern that are send to the different nodes of the hierarchy.}
\label{sm:fig_3}
\end{figure}

\begin{figure}[h]
\centering
\setlength\tabcolsep{1pt}
\begin{tabular}{@{}cc@{}}
 \includegraphics[width=0.49\linewidth]{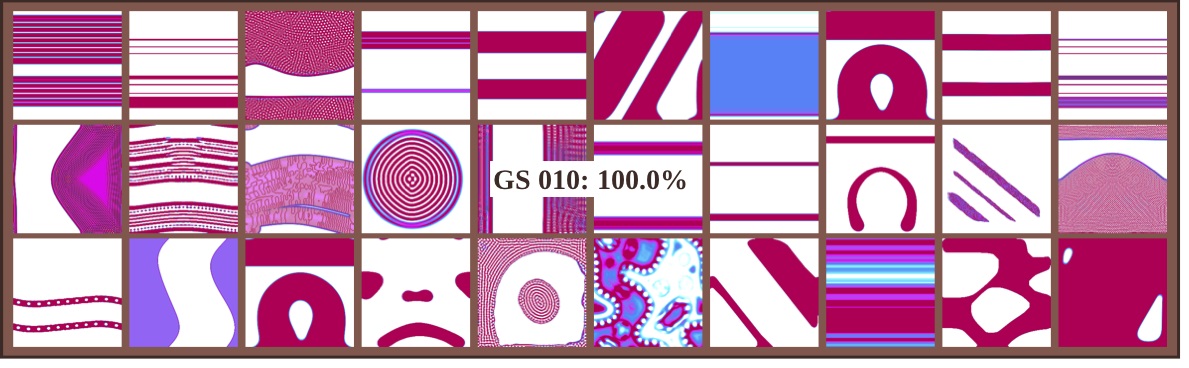} 
 & 
 \includegraphics[width=0.49\linewidth]{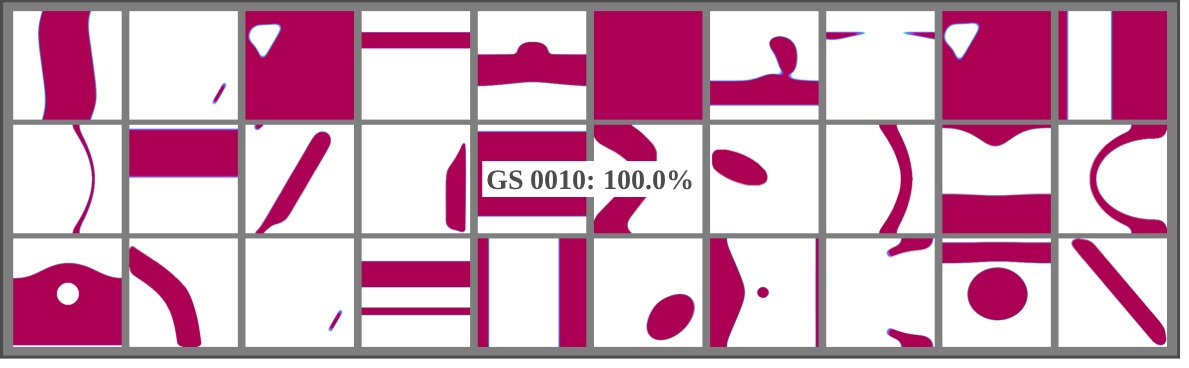}
\\
 \includegraphics[width=0.49\linewidth]{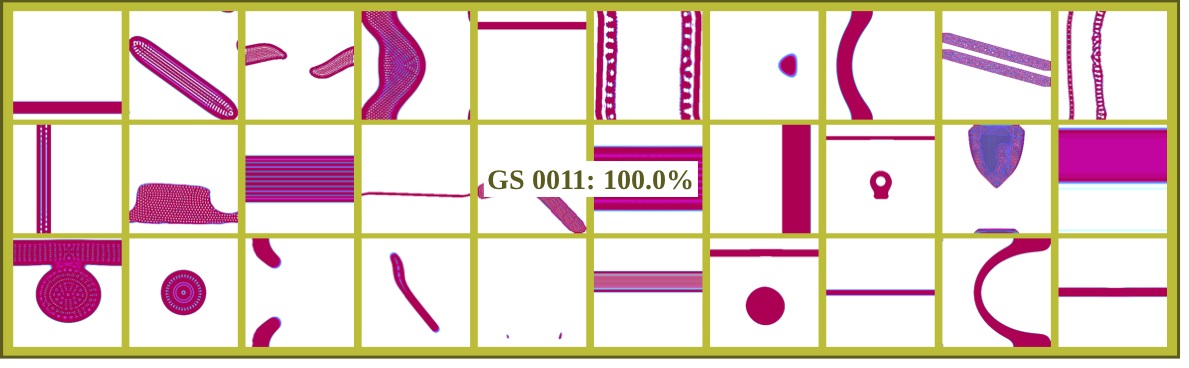}
 &
 \includegraphics[width=0.49\linewidth]{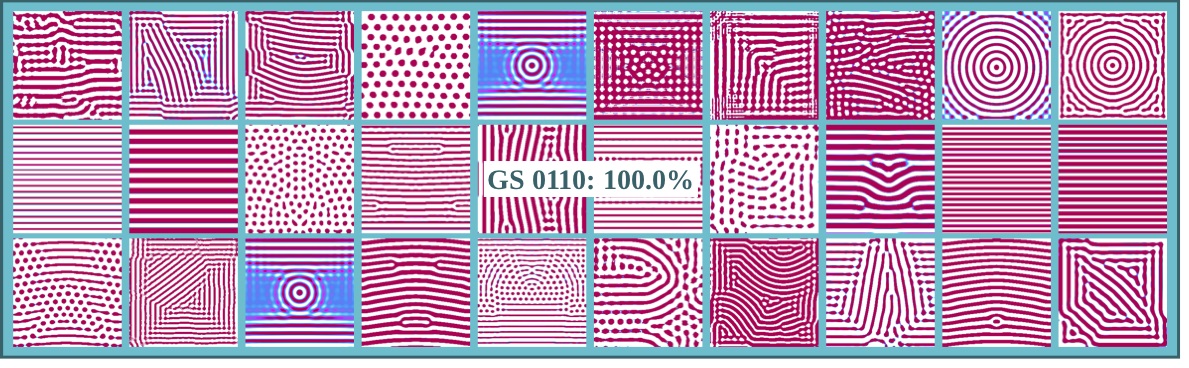}
 \\
 \includegraphics[width=0.49\linewidth]{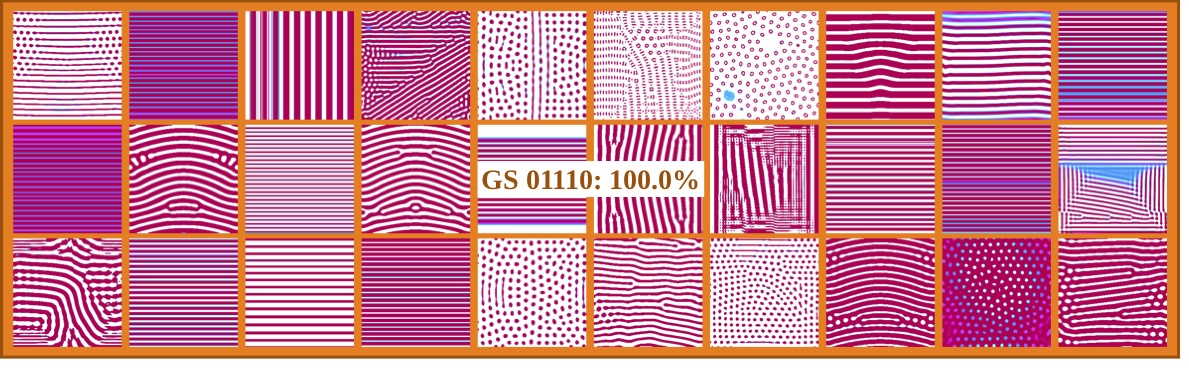}
 &
 \includegraphics[width=0.49\linewidth]{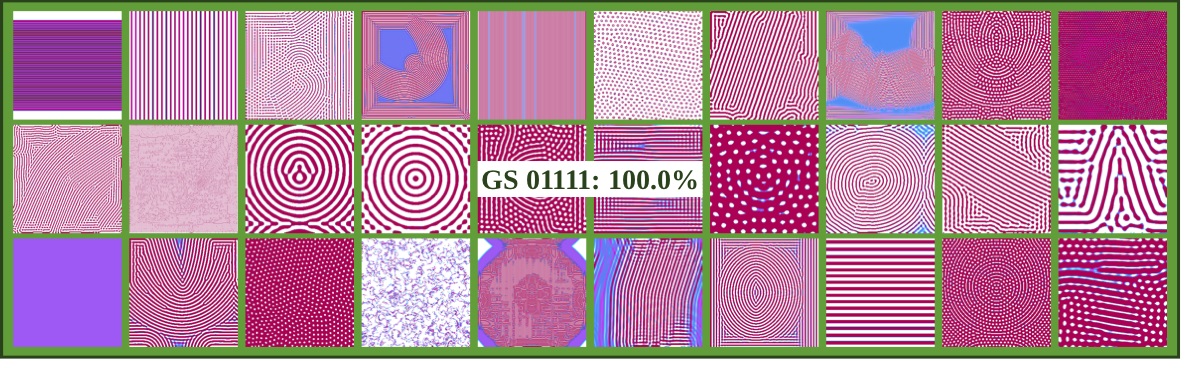}
\end{tabular}
 \caption{More example of discovered patterns in all leaf nodes (except GS 000 which gathers \enquote{dead} patterns).}
\label{sm:fig_4}
\end{figure}

\begin{figure}[h]
\begin{center}
\includegraphics[angle=0,width=1.1\linewidth]{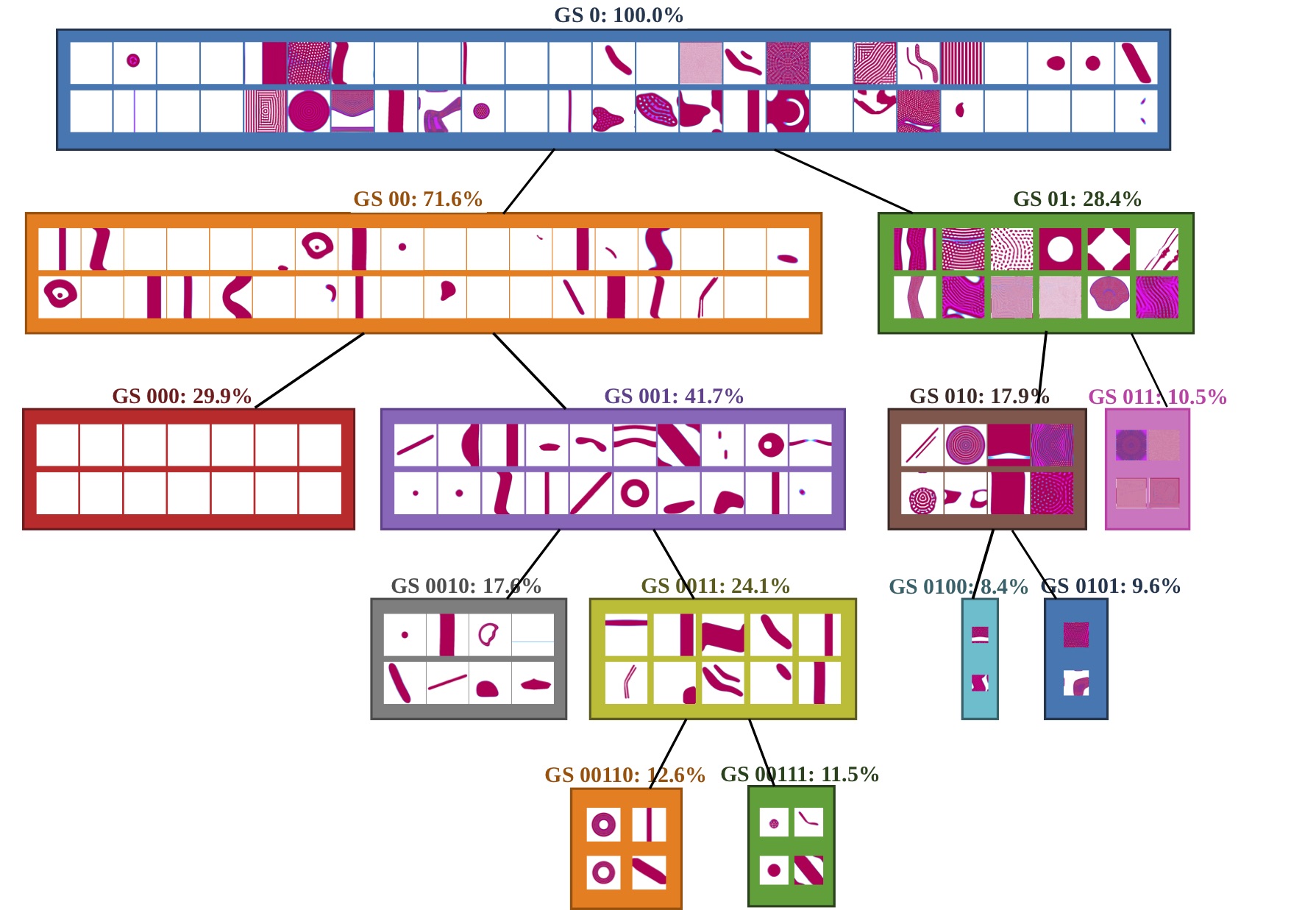}
\end{center}
\caption{Tree constructed by the IMGEP-HOLMES(A) algorithm during a single exploration with 5000 iterations. We display (randomly selected) discovered pattern that are send to the different nodes of the hierarchy.}
\label{sm:fig_5}
\end{figure}

\begin{figure}[h]
\centering
\setlength\tabcolsep{1pt}
\begin{tabular}{@{}cc@{}}
 \includegraphics[width=0.49\linewidth]{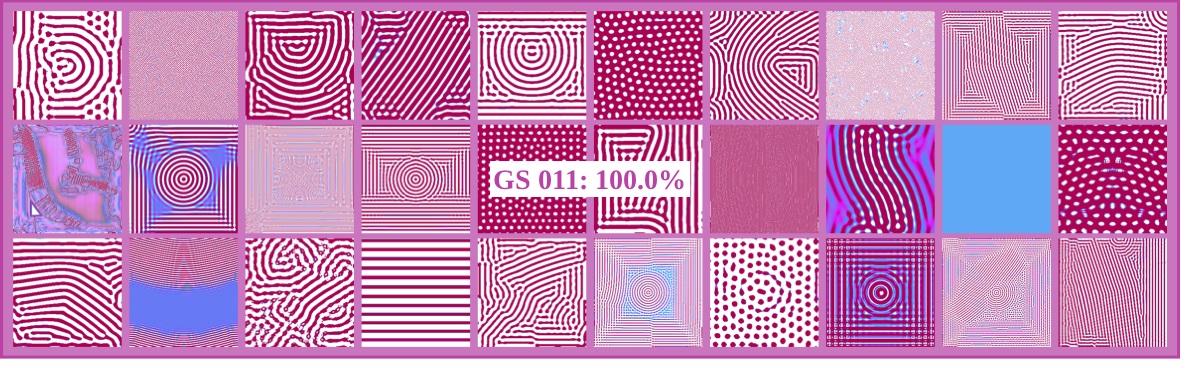} 
 & 
 \includegraphics[width=0.49\linewidth]{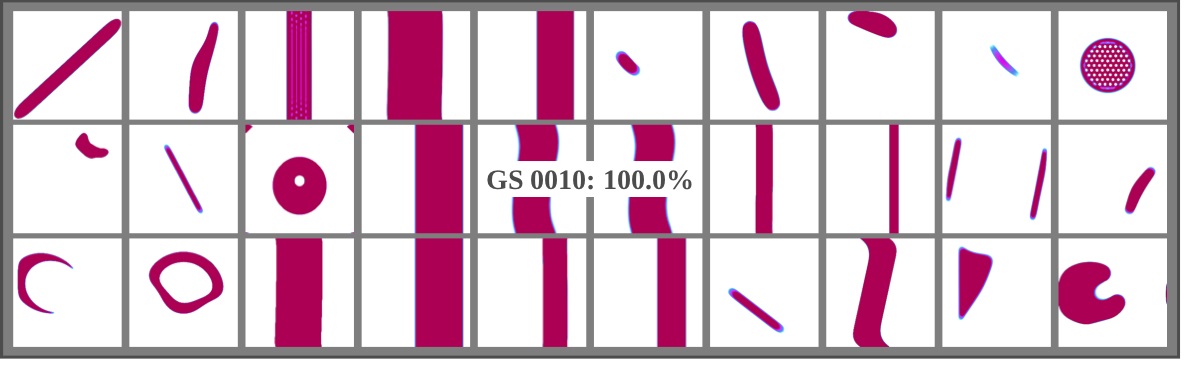}
\\
 \includegraphics[width=0.49\linewidth]{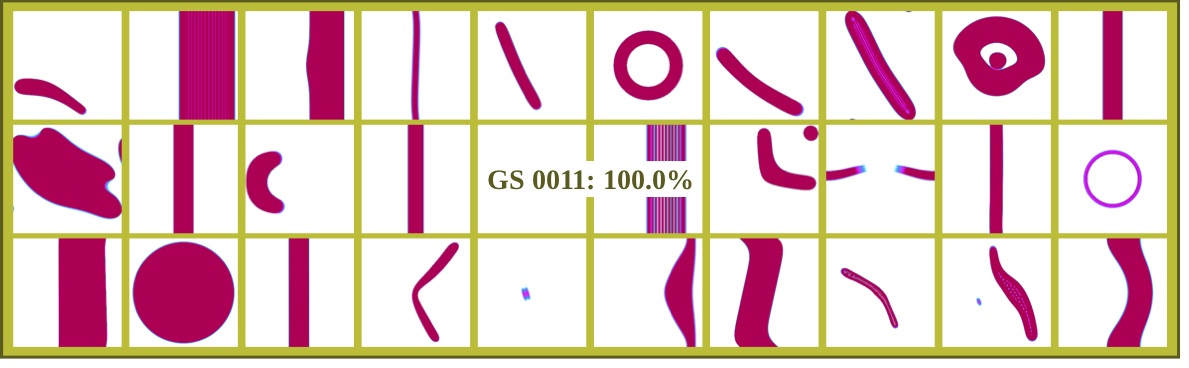}
 &
 \includegraphics[width=0.49\linewidth]{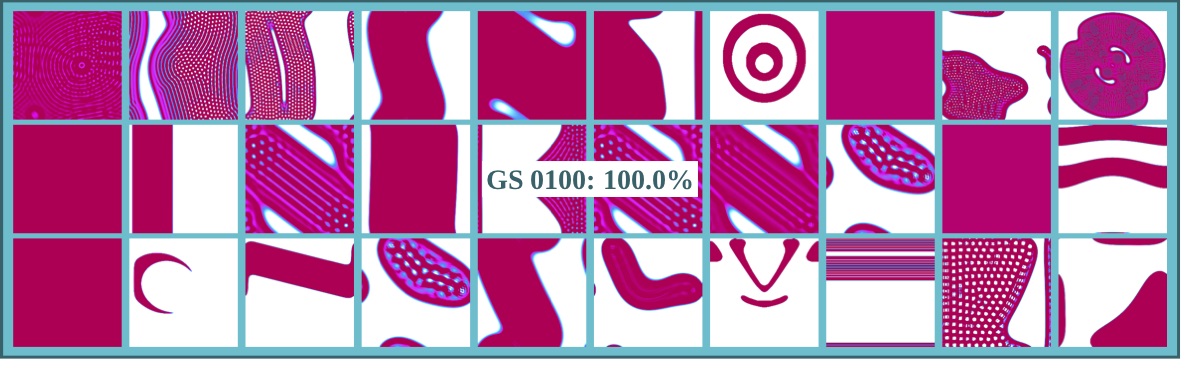}
 \\
 \includegraphics[width=0.49\linewidth]{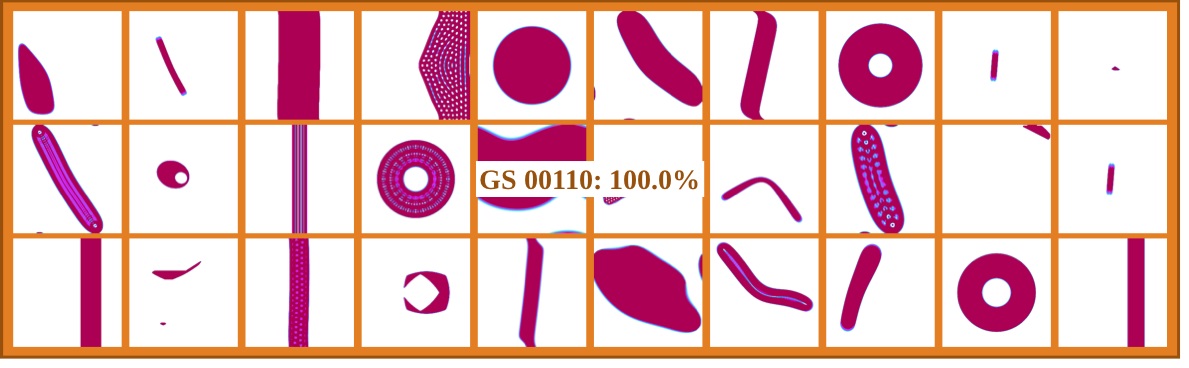}
 &
 \includegraphics[width=0.49\linewidth]{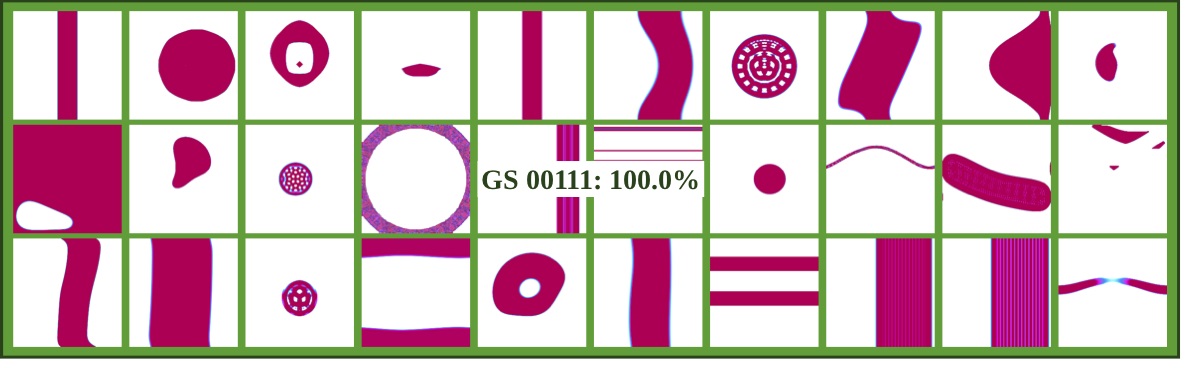}
\end{tabular}
 \caption{More example of discovered patterns in several leaf nodes.}
\label{sm:fig_6}
\end{figure}

\begin{figure}[h]
\begin{center}
\includegraphics[angle=0,width=0.75\linewidth]{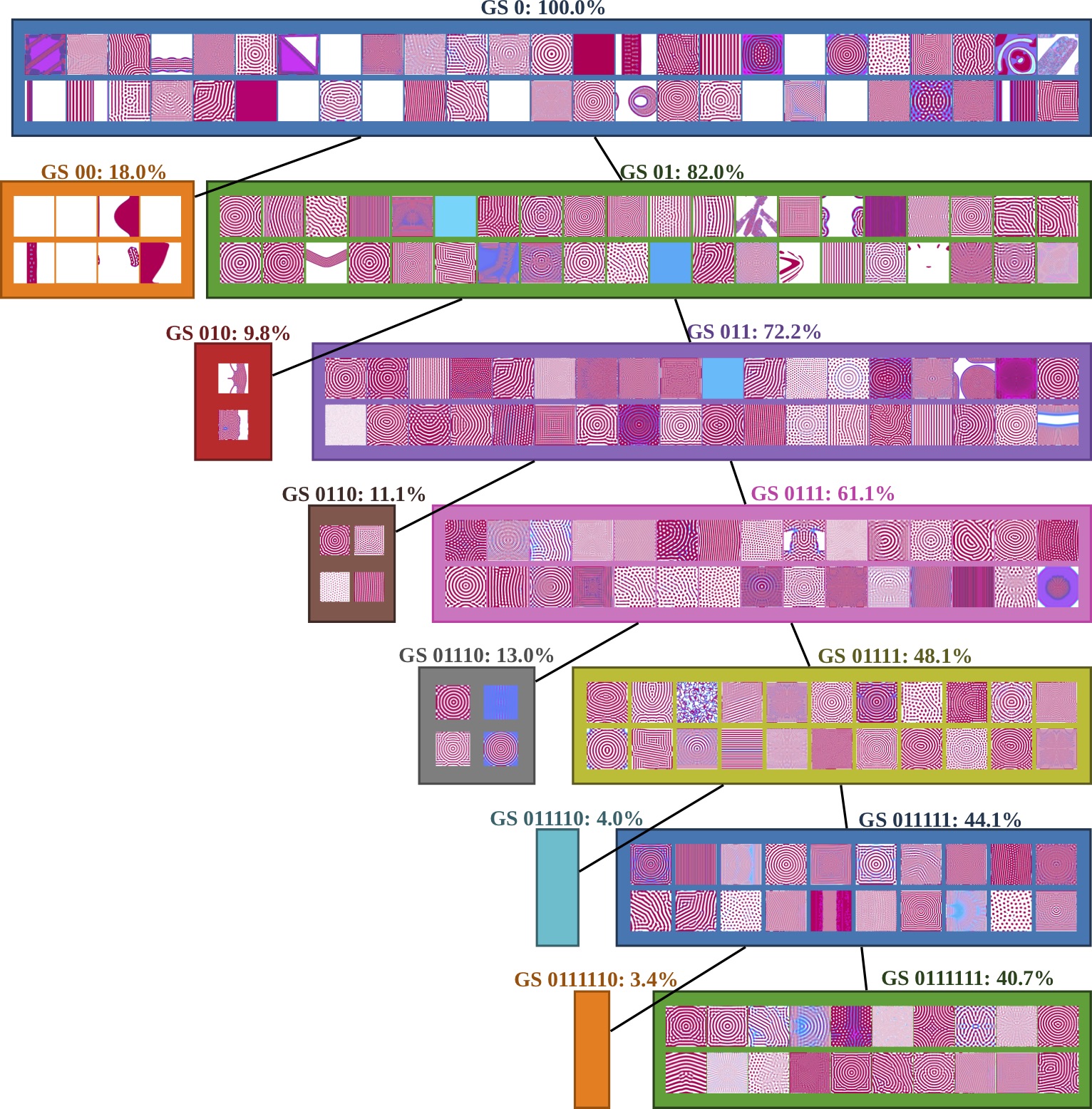}
\end{center}
\caption{Tree constructed by the IMGEP-HOLMES(NA) algorithm during a single exploration with 5000 iterations. We display (randomy selected) discovered pattern that are send to the different nodes of the hierarchy.}
\label{sm:fig_7}
\end{figure}

\begin{figure}[h]
\centering
\setlength\tabcolsep{1pt}
\begin{tabular}{@{}cc@{}}
 \includegraphics[width=0.49\linewidth]{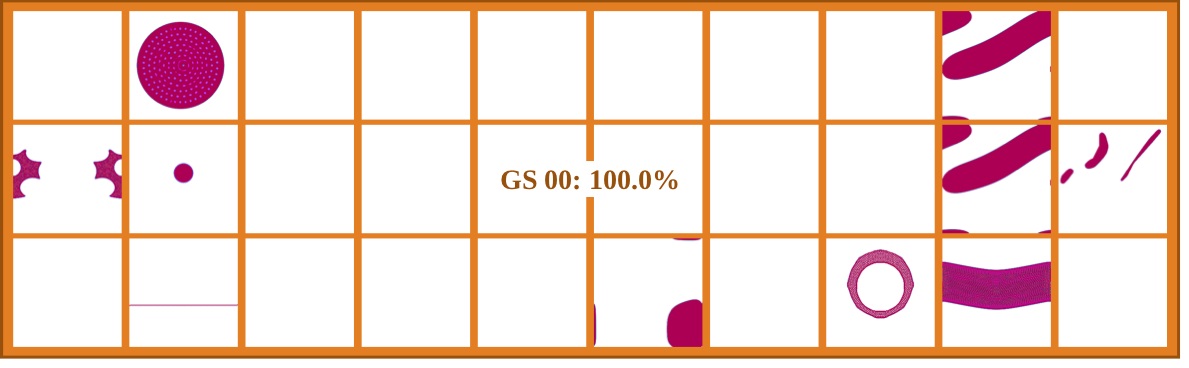} 
 & 
 \includegraphics[width=0.49\linewidth]{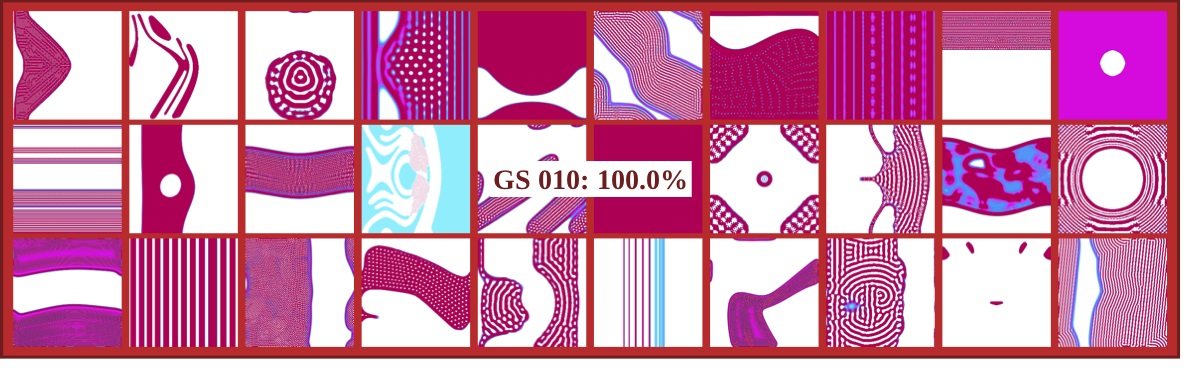}
\\
 \includegraphics[width=0.49\linewidth]{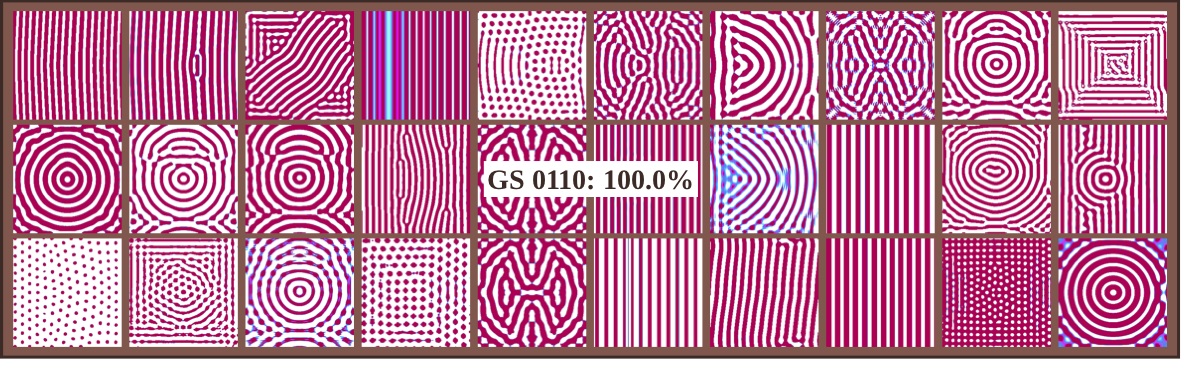}
 &
 \includegraphics[width=0.49\linewidth]{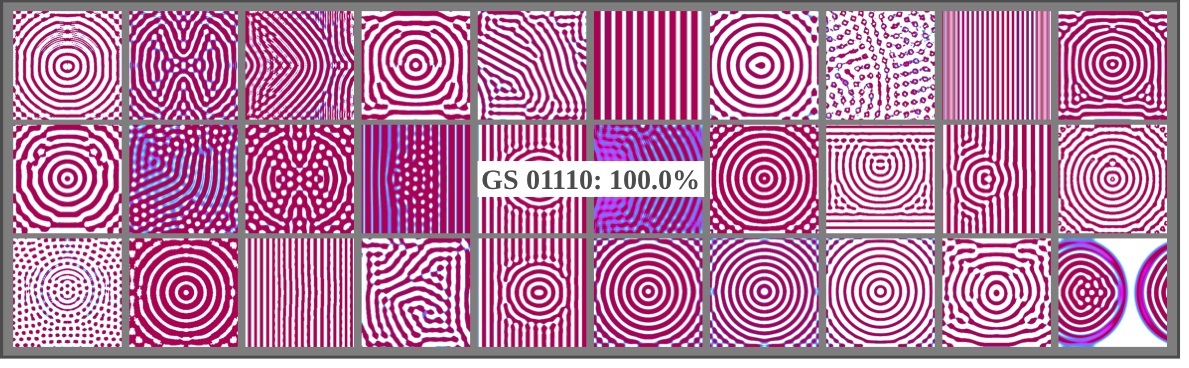}
 \\
 \includegraphics[width=0.49\linewidth]{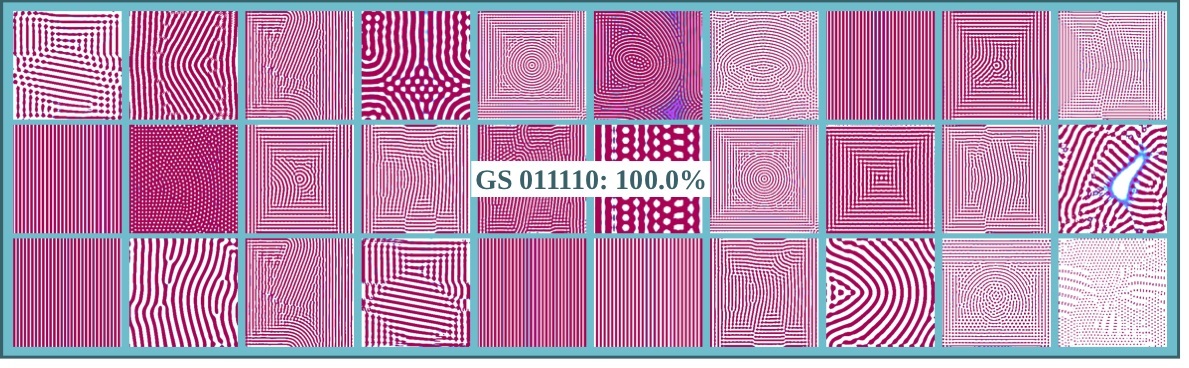}
 &
 \includegraphics[width=0.49\linewidth]{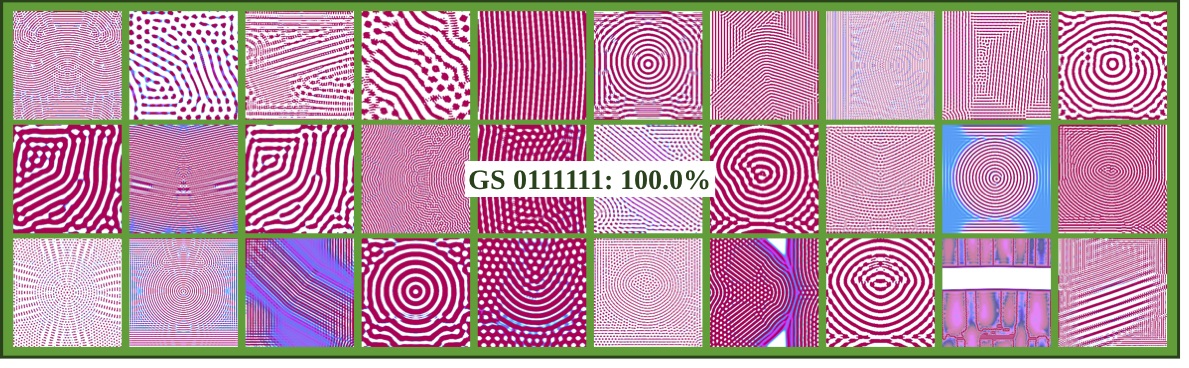}
\end{tabular}
 \caption{More example of discovered patterns in several leaf nodes.}
\label{sm:fig_8}
\end{figure}

The following visual results enable a more intuitive interpretation of the quantitative results presented in the main paper. Figure \ref{sm:fig_3}, \ref{sm:fig_5} and \ref{sm:fig_7} illustrate the final hierarchical tree that has been incrementally created by the different IMGEP-HOLMES variants. We can see how the discoverted patterns are partitionned along the hierarchy. Figure \ref{sm:fig_4}, \ref{sm:fig_6} and \ref{sm:fig_8} show additional illustration of patterns (randomly selected) in the final leaves of the tree. These figures illustrate how exploration guidance can drive the type of found diversity. For instance, we can see that IMGEP-HOLMES(A) allocates more goal space nodes for \enquote{animal} patterns whereas IMGEP-HOLMES(NA) discovers majoritary \enquote{non-animal} patterns.\\
Additionally, section \ref{sm:sec_A_1_4} provides examples of patterns reconstructed by HOLMES representation showing the coarse-to-fine specialisation along the tree.
\clearpage
\subsubsection{IMGEP-HOLMES discoveries}
\label{sm:sec_A_1_1}

\clearpage
\subsubsection{IMGEP-HOLMES(A) discoveries}
\label{sm:sec_A_1_2}

\clearpage
\subsubsection{IMGEP-HOLMES(NA) discoveries}
\label{sm:sec_A_1_3}

\clearpage
\subsubsection{Coarse-to-fine specialisation}
\label{sm:sec_A_1_4}
\begin{figure}[h!]
\centering
\setlength\tabcolsep{2pt}
\begin{tabular}{c@{\hskip 5pt}|@{\hskip 5pt}ccccc}
\multirow{2}{*}{Original image}& \multicolumn{5}{c}{HOLMES reconstructions} \\
& depth 0 & depth 1 & depth 2 & depth 3 & depth 4 \\
 \fbox{\includegraphics[width=0.07\linewidth]{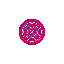}}
 & 
 \fbox{\includegraphics[width=0.07\linewidth]{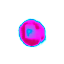}}
 &
 \fbox{\includegraphics[width=0.07\linewidth]{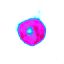}}
 &
 \fbox{\includegraphics[width=0.07\linewidth]{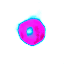}}
 &
 \fbox{\includegraphics[width=0.07\linewidth]{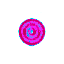}}
 &
 
 \\[4pt]
 
 \fbox{\includegraphics[width=0.07\linewidth]{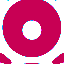}}
  &
 \fbox{\includegraphics[width=0.07\linewidth]{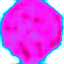}}
 &
 \fbox{\includegraphics[width=0.07\linewidth]{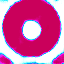}}
 &
 \fbox{\includegraphics[width=0.07\linewidth]{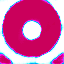}}
 &

 &
 
 \\[4pt]

 \fbox{\includegraphics[width=0.07\linewidth]{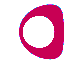}}
 & 
 \fbox{\includegraphics[width=0.07\linewidth]{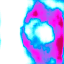}}
 &
 \fbox{\includegraphics[width=0.07\linewidth]{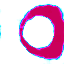}}
 &
 \fbox{\includegraphics[width=0.07\linewidth]{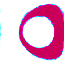}}
 &

 &
 
 \\[4pt]

 \fbox{\includegraphics[width=0.07\linewidth]{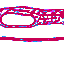}}
 & 
 \fbox{\includegraphics[width=0.07\linewidth]{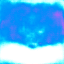}}
 &
 \fbox{\includegraphics[width=0.07\linewidth]{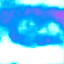}}
 &
 \fbox{\includegraphics[width=0.07\linewidth]{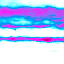}}
 &

 &
 
 \\[4pt]

 \fbox{\includegraphics[width=0.07\linewidth]{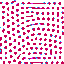}}
 & 
 \fbox{\includegraphics[width=0.07\linewidth]{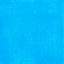}}
 &
 \fbox{\includegraphics[width=0.07\linewidth]{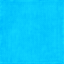}}
 &
 \fbox{\includegraphics[width=0.07\linewidth]{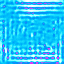}}
 &
 \fbox{\includegraphics[width=0.07\linewidth]{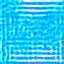}}
 &
 
 \\[4pt]

 \fbox{\includegraphics[width=0.07\linewidth]{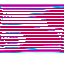}}
 & 
 \fbox{\includegraphics[width=0.07\linewidth]{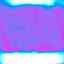}}
 &
 \fbox{\includegraphics[width=0.07\linewidth]{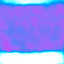}}
 &
 \fbox{\includegraphics[width=0.07\linewidth]{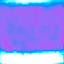}}
 &
 \fbox{\includegraphics[width=0.07\linewidth]{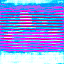}}
 &
 \fbox{\includegraphics[width=0.07\linewidth]{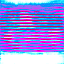}}
 \\[4pt]

 \fbox{\includegraphics[width=0.07\linewidth]{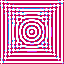}} 
 & 
 \fbox{\includegraphics[width=0.07\linewidth]{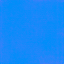}}
 &
 \fbox{\includegraphics[width=0.07\linewidth]{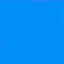}}
 &
 \fbox{\includegraphics[width=0.07\linewidth]{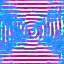}}
 &
 \fbox{\includegraphics[width=0.07\linewidth]{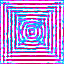}}
 &
\end{tabular}
 \caption{Examples of patterns and their reconstructions along HOLMES tree. Please note that all patterns are originally gray-scale and that, for visualisation purpose, we follow the color scheme of \cite{reinke2019intrinsically}.}
\label{sm:fig_11}
\end{figure}

This section provides the reconstruction performances of HOLMES representation, learned during the IMGEP-HOLMES experiment, on an external test dataset of 750 images. The results are summarized in table \ref{sm:table_1}. Figure \ref{sm:fig_11} provides additional examples of patterns and their reconstructions. We can see that HOLMES progressively learns to reconstruct more and more fine-grained details, which is a good proxy evaluation of HOLMES ability to learn coarse-to-fine representations.

\begin{table}[h!]
\caption{Reconstruction error, measured by pixel-wise binary cross entropy loss (BCE), on the test dataset. We provide mean and standard deviation over the different repetitions (n=5).}
\label{sm:table_1}
\begin{center}
\vspace{-2pt}
\begin{tabular}{lcc}
 & IMGEP-HOLMES root node representation & IMGEP-HOLMES leaf node  representation
\\ \hline 
\bf BCE &  19710 $\pm$ 722 & 17383 $\pm$ 301 $\mathbf{(\downarrow 2327)}$  \\
\end{tabular}
\end{center}
\end{table}

\newpage
\subsection{Statistical Results}
\label{sm:sec_A_2}

\subsubsection{Representational Similarity Analysis}

\begin{figure}[h!]
\begin{center}
\includegraphics[width=\linewidth]{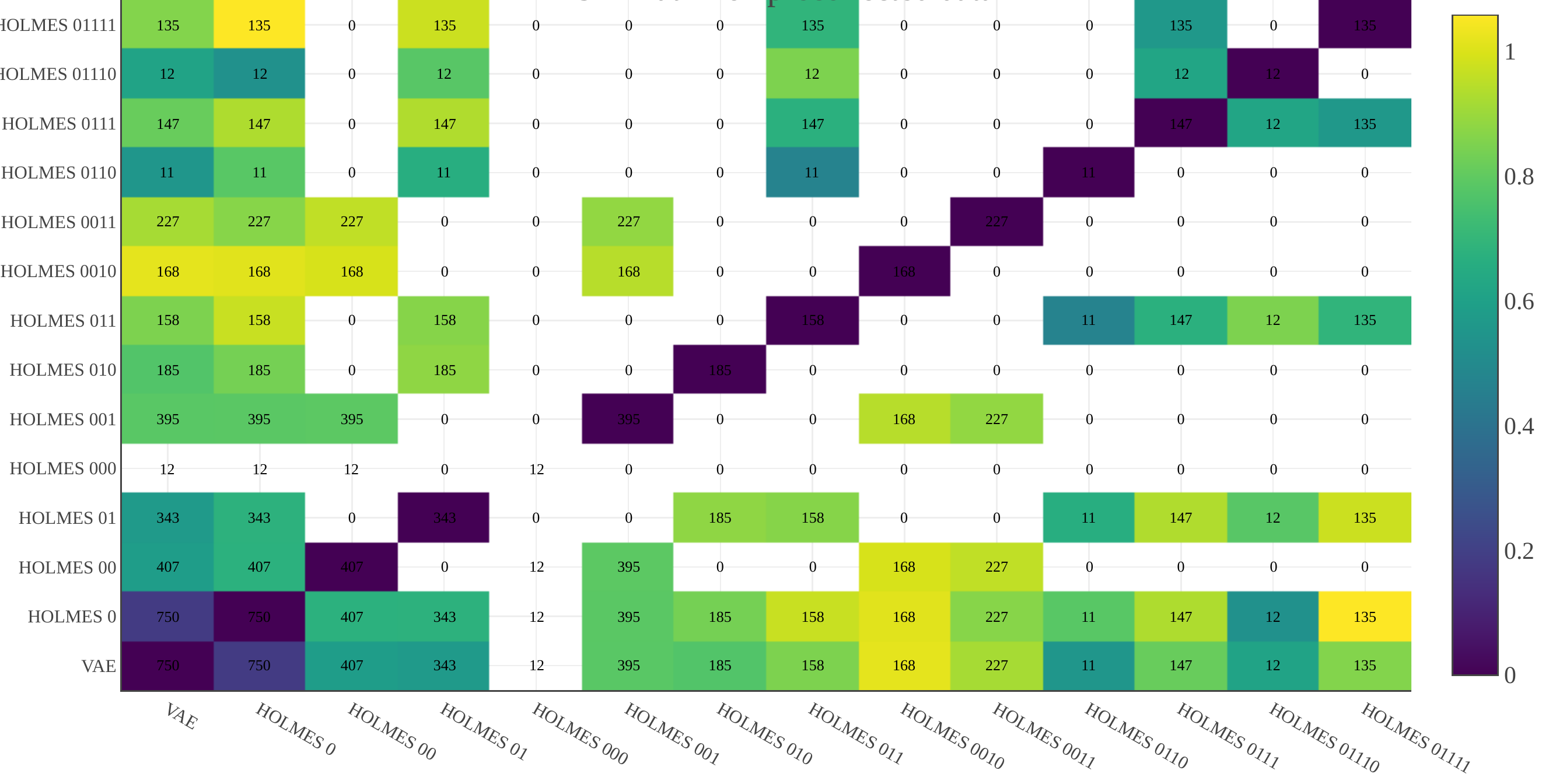}
\end{center}
\caption{RSA heatmap showing Spearman's $\rho$ correlation (colorscale) between disagreement among the different goal space representations. The displayed numbers represent the count of Lenia patterns (over the 750 patterns from an external precollected dataset) that are shared between the respective pair of goal spaces and based on which the dissimilarity was computed.}
\label{sm:fig_9}
\end{figure}
To evaluate the diversity of the representations achieved by the VAE and HOLMES architecture the representational similarity matrix has been calculated (Figure \ref{sm:fig_9}). Both VAE and HOLMES encode a set of pre-selected images (unbiased by exploration) and the achieved representations are compared by using the Spearman's $\rho$ correlation measure. Since VAE has only one goal space and HOLMES has one per node of the hierarchy, each goal space is mutually compared. Additionally, since HOLMES redirects images through the hierarchy only the images which are in common to both compared goal spaces have been used. The goal spaces which have no images in common are marked with the value 0 in the table.   

The dissimilarity index of two compared representations $R_n \in G_n $ and $R_m \in G_m$ is calculated in two stages. First, correlation distance of all the image representation pairs $x = [z_i,z_j]$  is calculated as follows as a dissimilarity measure. 
\begin{equation*}
    corr(x) = 1- \frac{(z_i - \overline{z}_i)\cdot(z_j - \overline{z}_j)}{||z_i - \overline{z}_i||_2||z_j - \overline{z}_j||_2}
\end{equation*}
The $\overline{z}_i$ is a mean value of $z_i$ elements, $\cdot$ is a dot product and $||.||_2$ is the $l_2$ norm. The result of this step is a $N \times N$ sized matrix for each representation $R_n$ and $R_m$, where $N$ is the number of images in common, showing the correlation distance of each image pairs in its goal space. Second phase is calculation of the Spearman's $\rho$ rank correlation coefficient. The Spearman's coefficient is a standard statistical method for determining the significance of correlation between two data sets where $\rho \in [0,1]$. The closer the value of $\rho$ to 1 the higher the correlation in representations. In the figure \ref{sm:fig_9} the $\rho$ coefficient is depicted by using a heatmap colors and the displayed numbers indicate the number of images in common for each representation pair.

\subsubsection{Diversity explored by the IMGEP variant in the different goal space representations}
\begin{figure}[ht!]
\centering
\resizebox{10cm}{!}{%
\setlength\tabcolsep{1pt}
\begin{tabular}{@{}cc@{}}
 Diversity in VAE Goal Space  & Diversity in HOLMES GS 0 \\
 \includegraphics[width=0.4\linewidth]{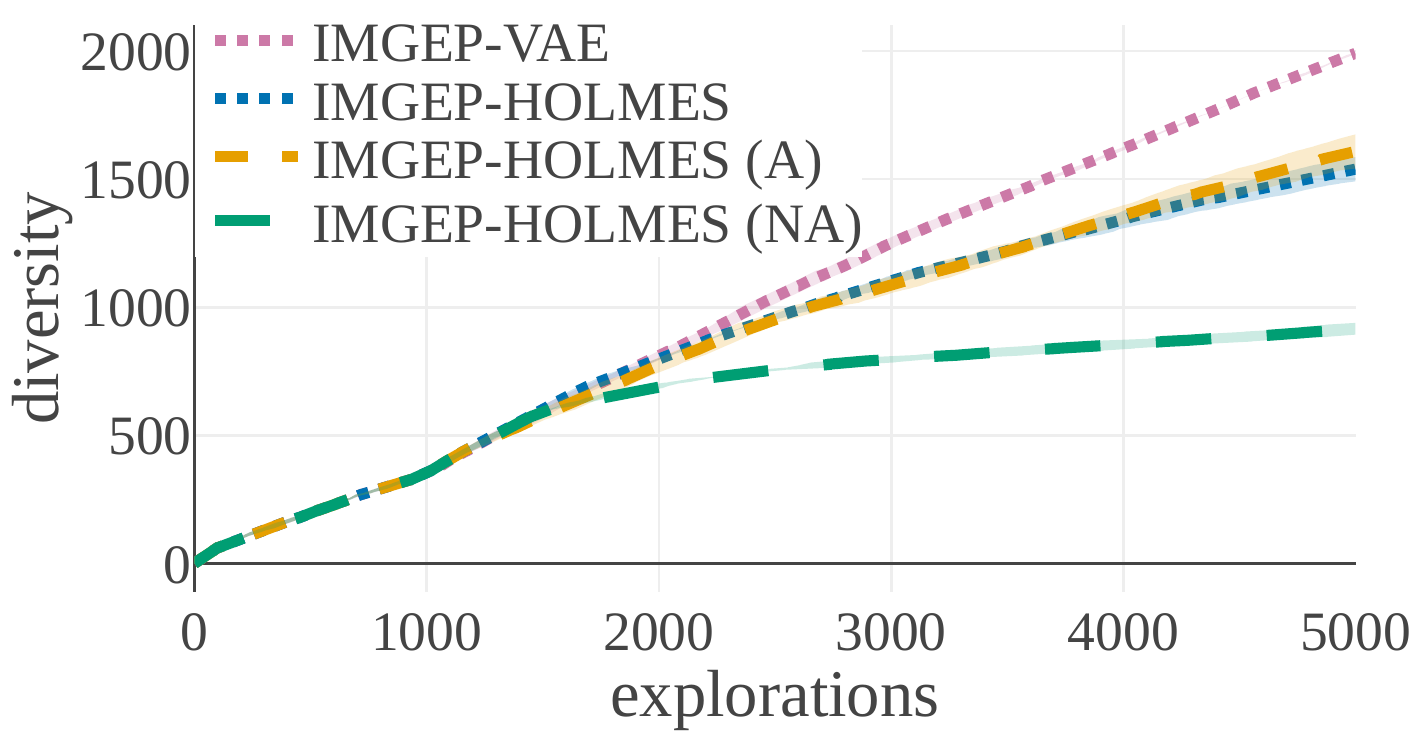}
 &
 \includegraphics[width=0.4\linewidth]{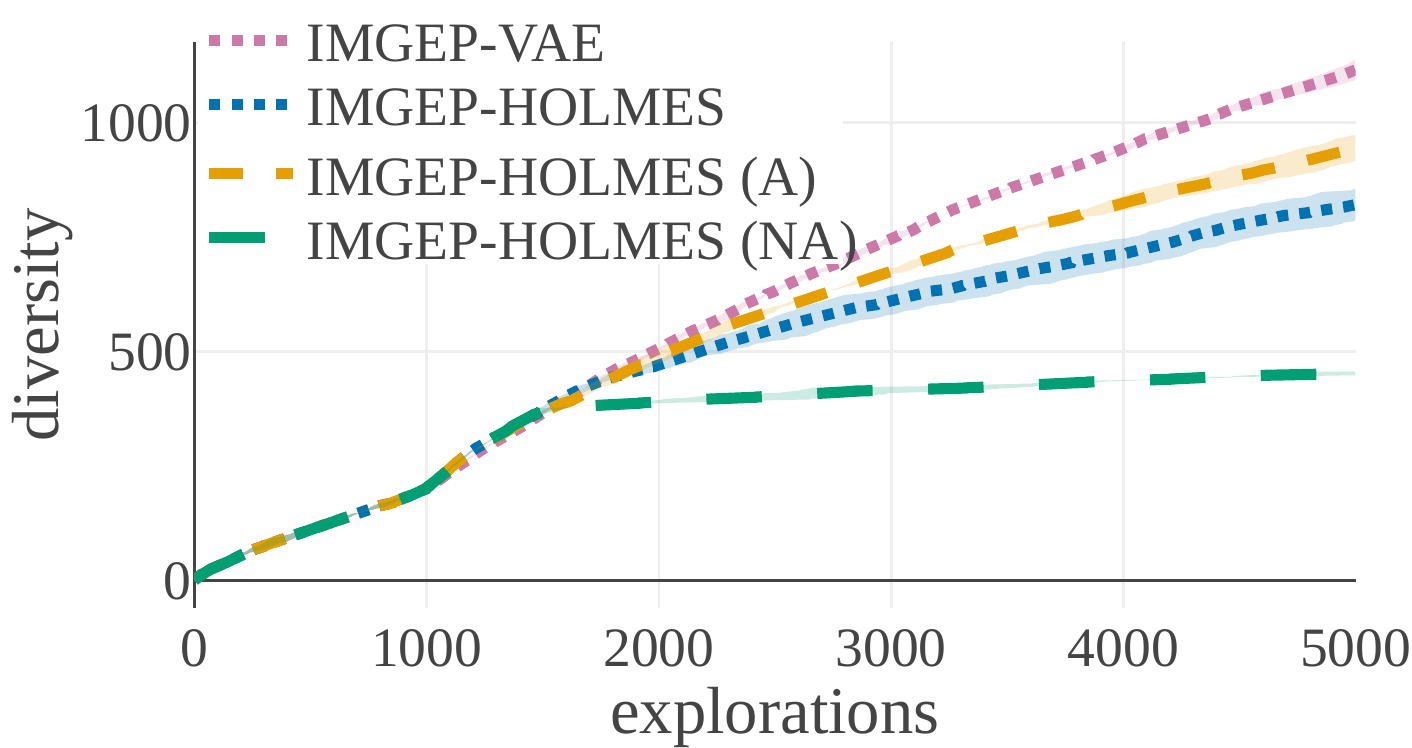} \\
 
  Diversity in HOLMES GS 00  & Diversity in HOLMES GS 01 \\
 \includegraphics[width=0.4\linewidth]{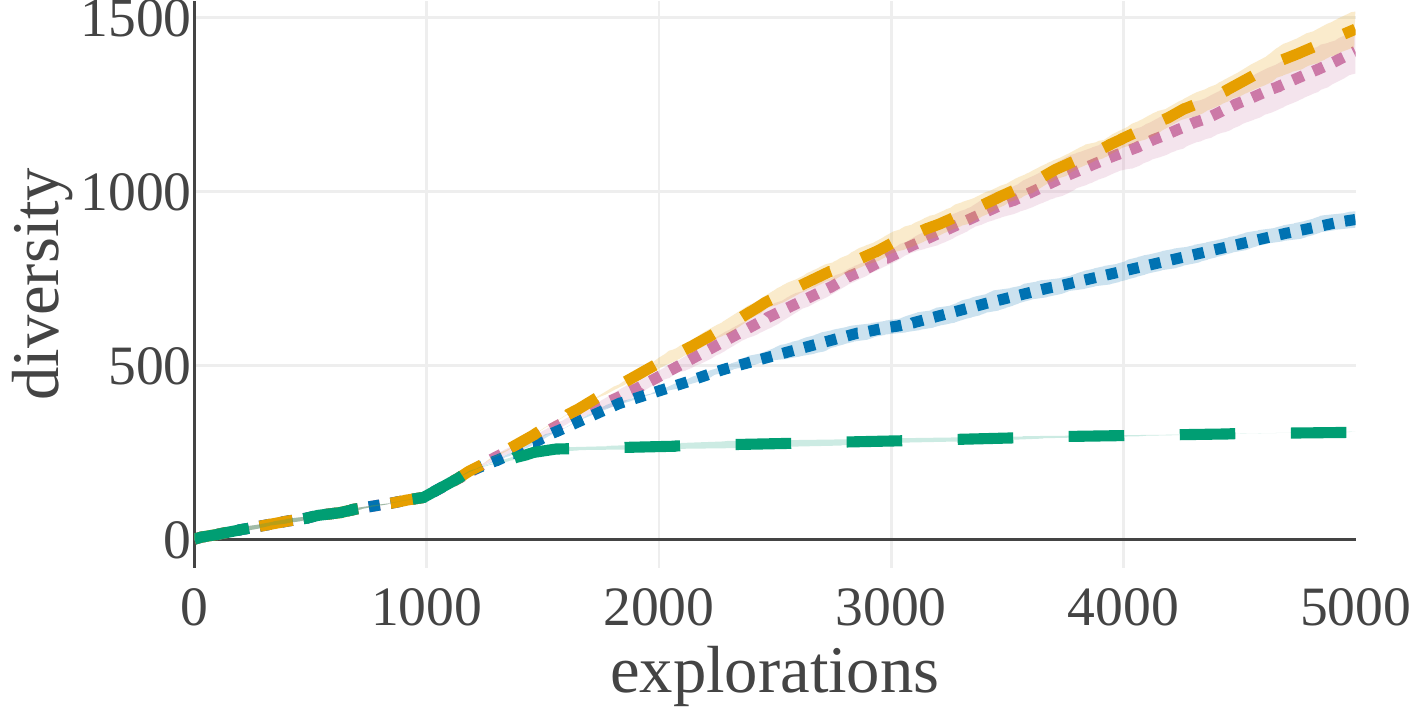} 
 &
 \includegraphics[width=0.4\linewidth]{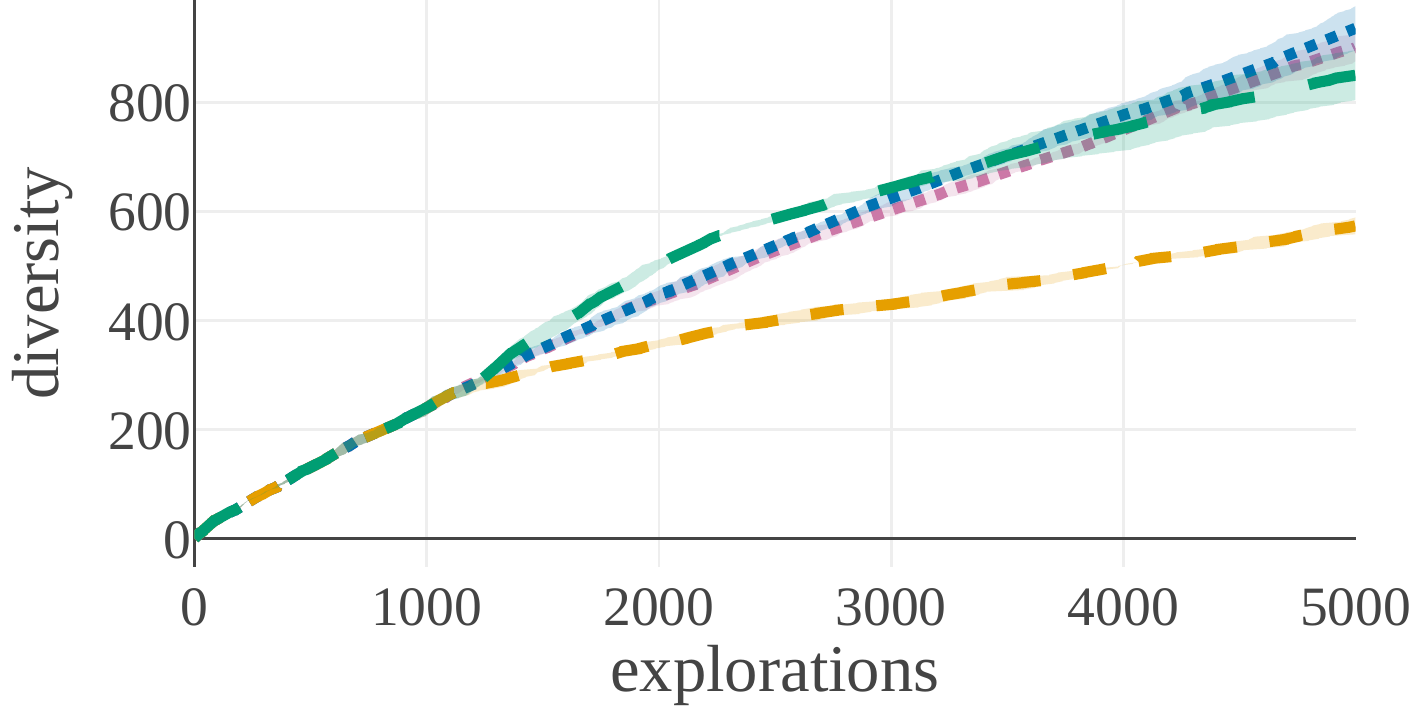} \\
 
  Diversity in HOLMES GS 000 & Diversity in HOLMES GS 001 \\
 \includegraphics[width=0.4\linewidth]{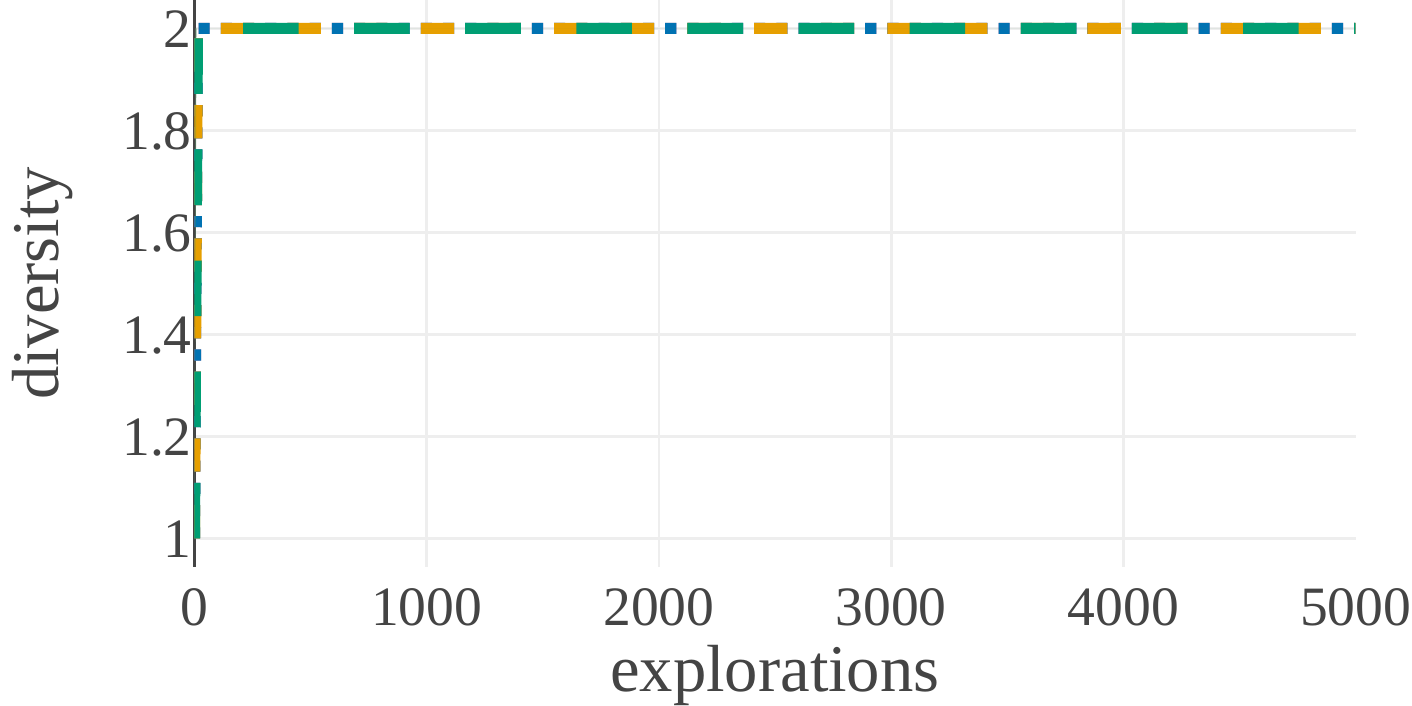} 
 &
 \includegraphics[width=0.4\linewidth]{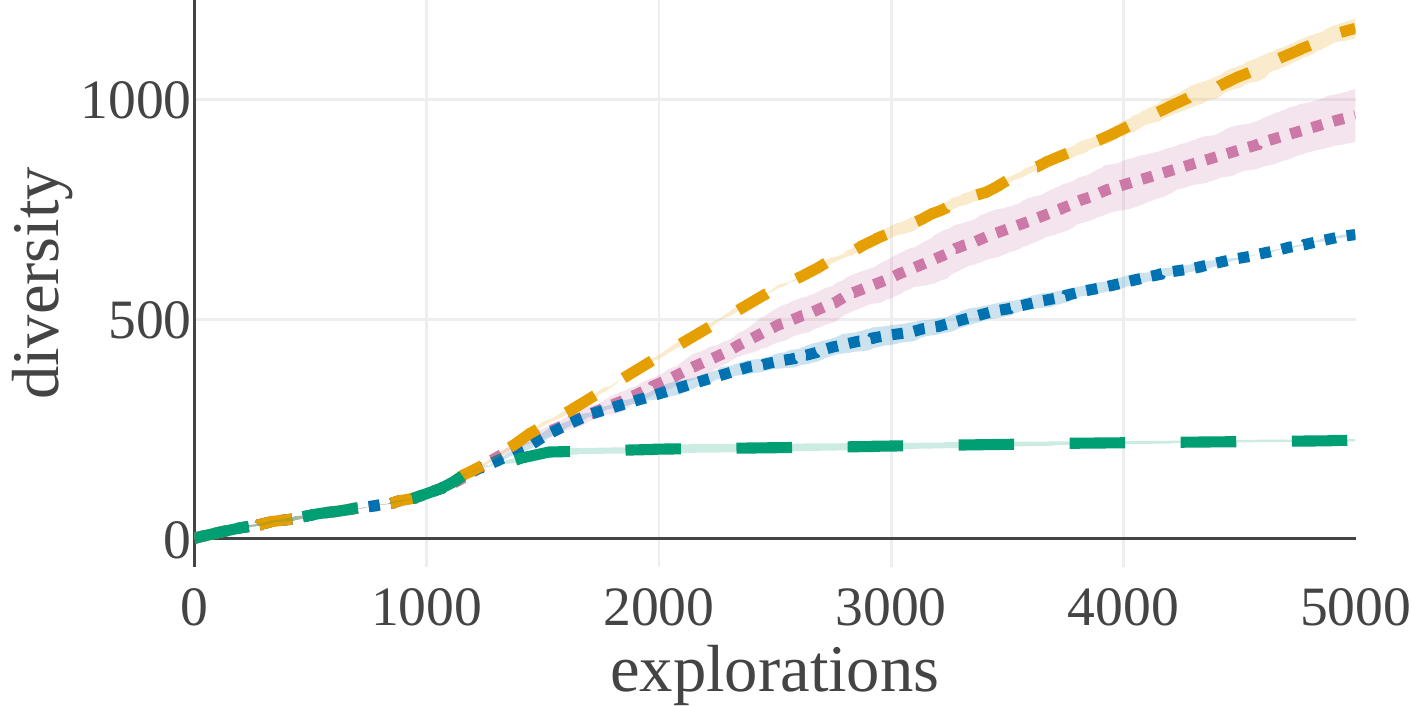} \\
 
  Diversity in HOLMES GS 010  & Diversity in HOLMES GS 011 \\
 \includegraphics[width=0.4\linewidth]{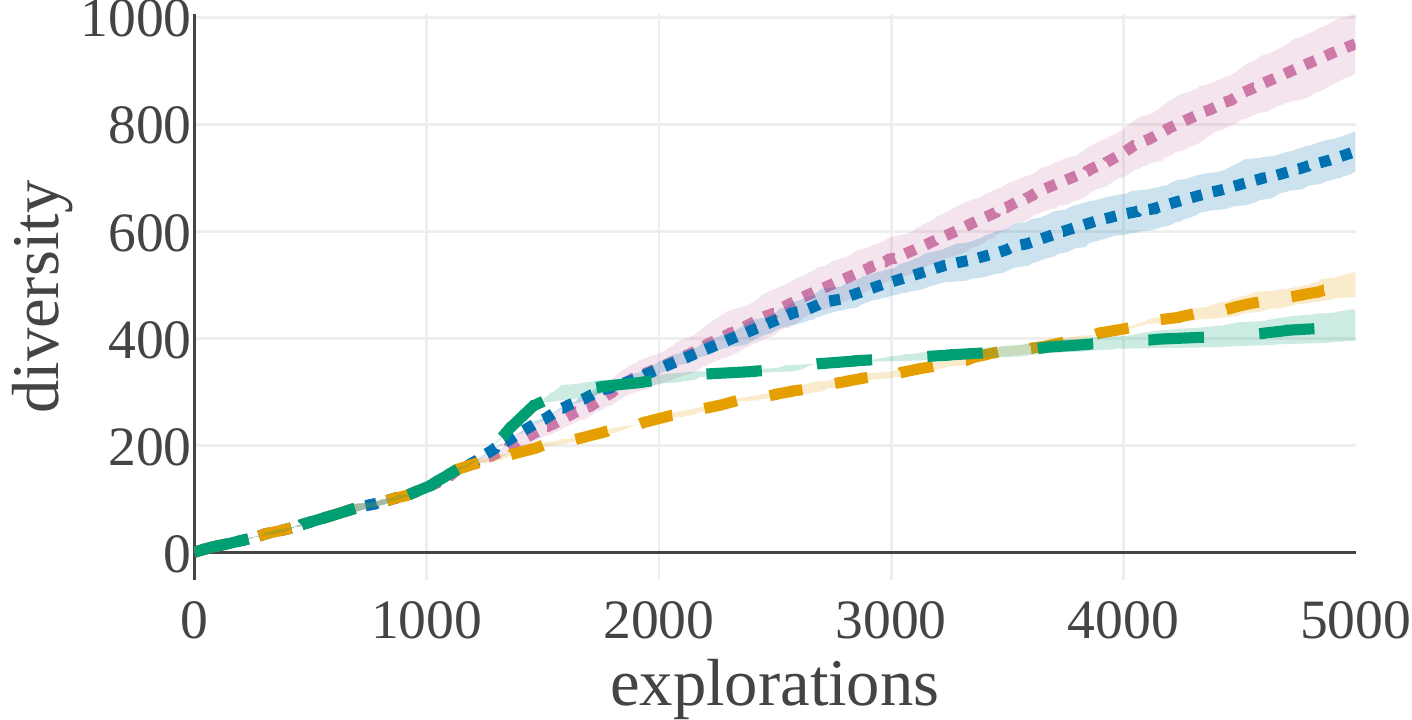} 
 &
 \includegraphics[width=0.4\linewidth]{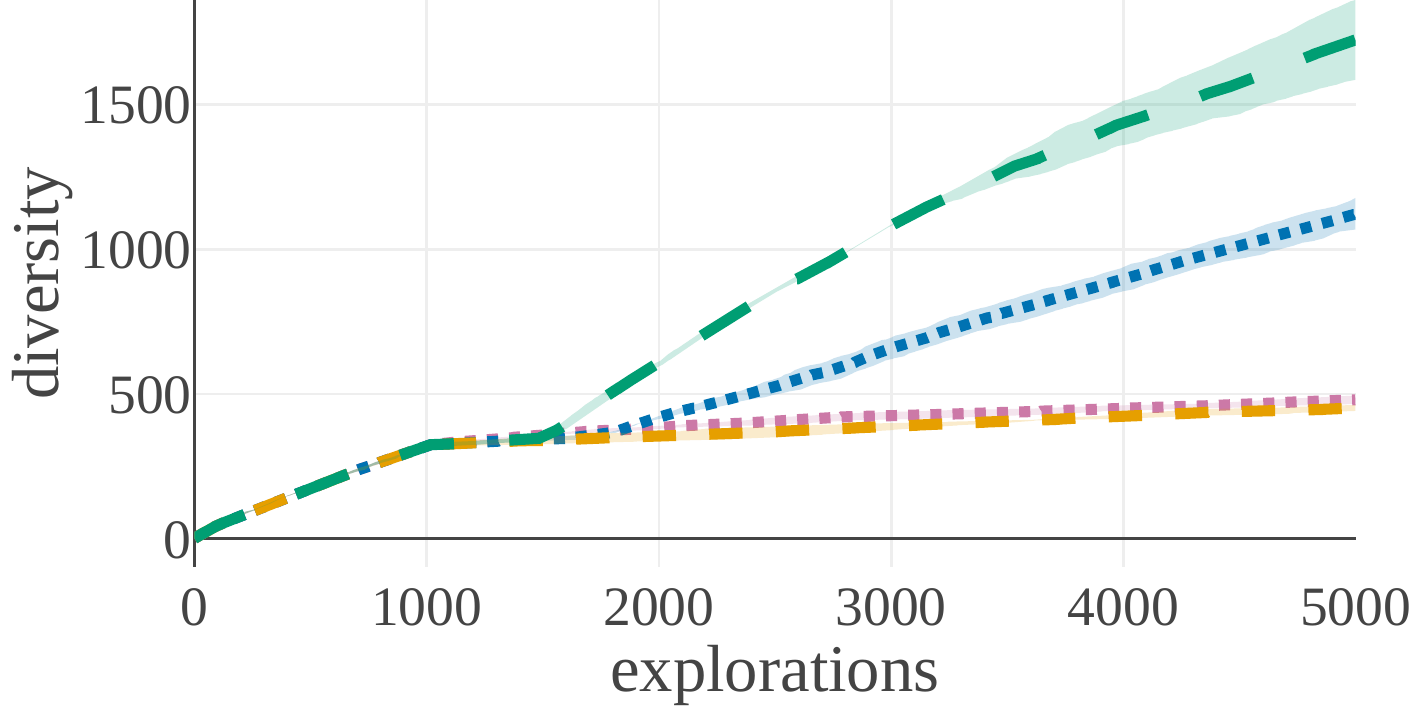} \\
 
   Diversity in HOLMES GS 0010 & Diversity in HOLMES GS 0011 \\
 \includegraphics[width=0.4\linewidth]{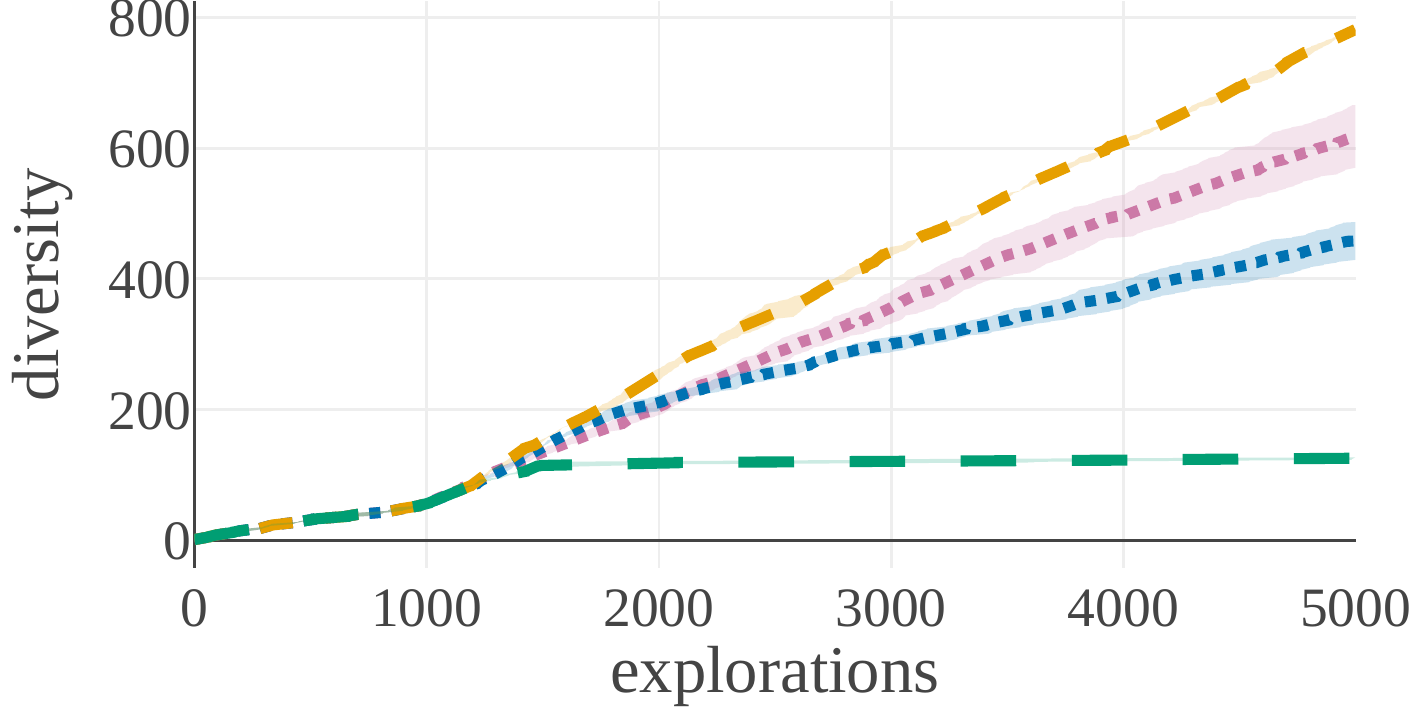} 
 &
\includegraphics[width=0.4\linewidth]{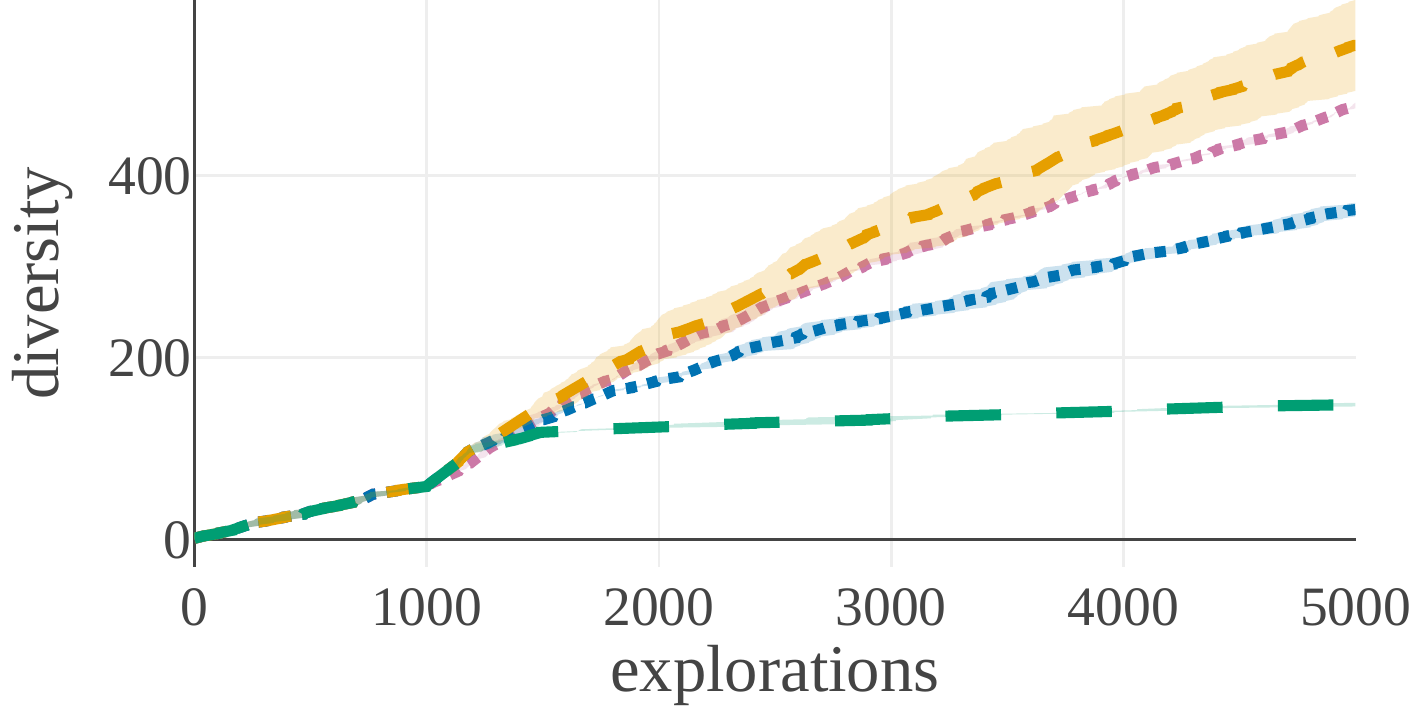} \\
 
  Diversity in HOLMES GS 0110  & Diversity in HOLMES GS 0111 \\
 \includegraphics[width=0.4\linewidth]{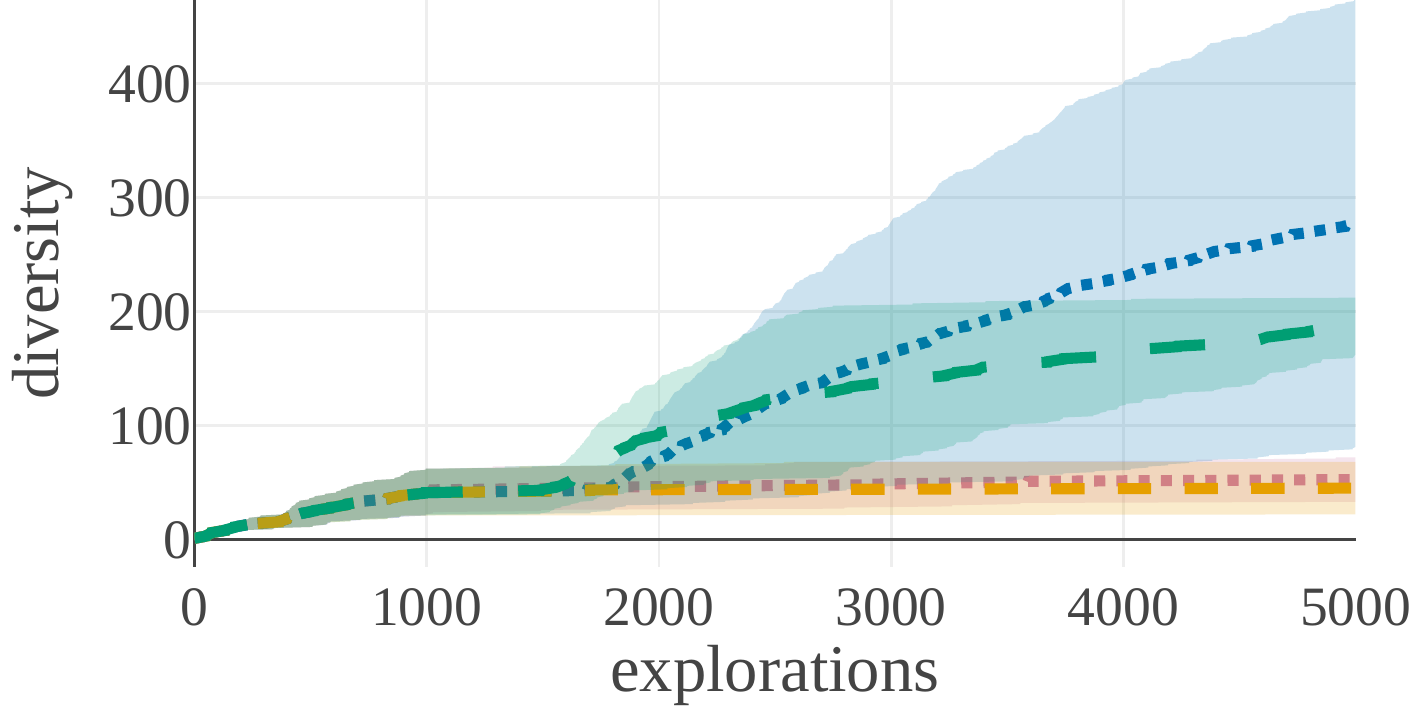} 
 &
 \includegraphics[width=0.4\linewidth]{diversity_holmesgoalspace_011_all_aranged.pdf} \\
 
   Diversity in HOLMES GS 01110  & Diversity in HOLMES GS 01111 \\
 \includegraphics[width=0.4\linewidth]{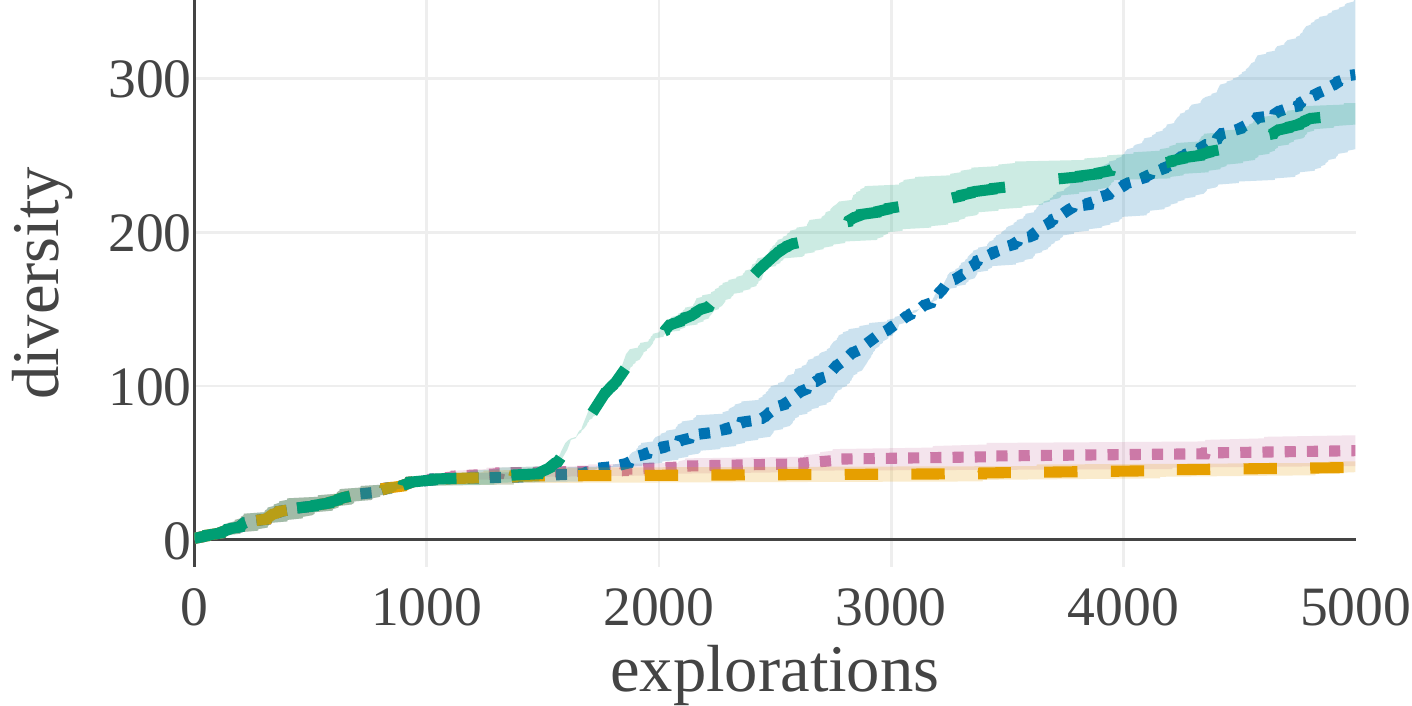} 
 &
 \includegraphics[width=0.4\linewidth]{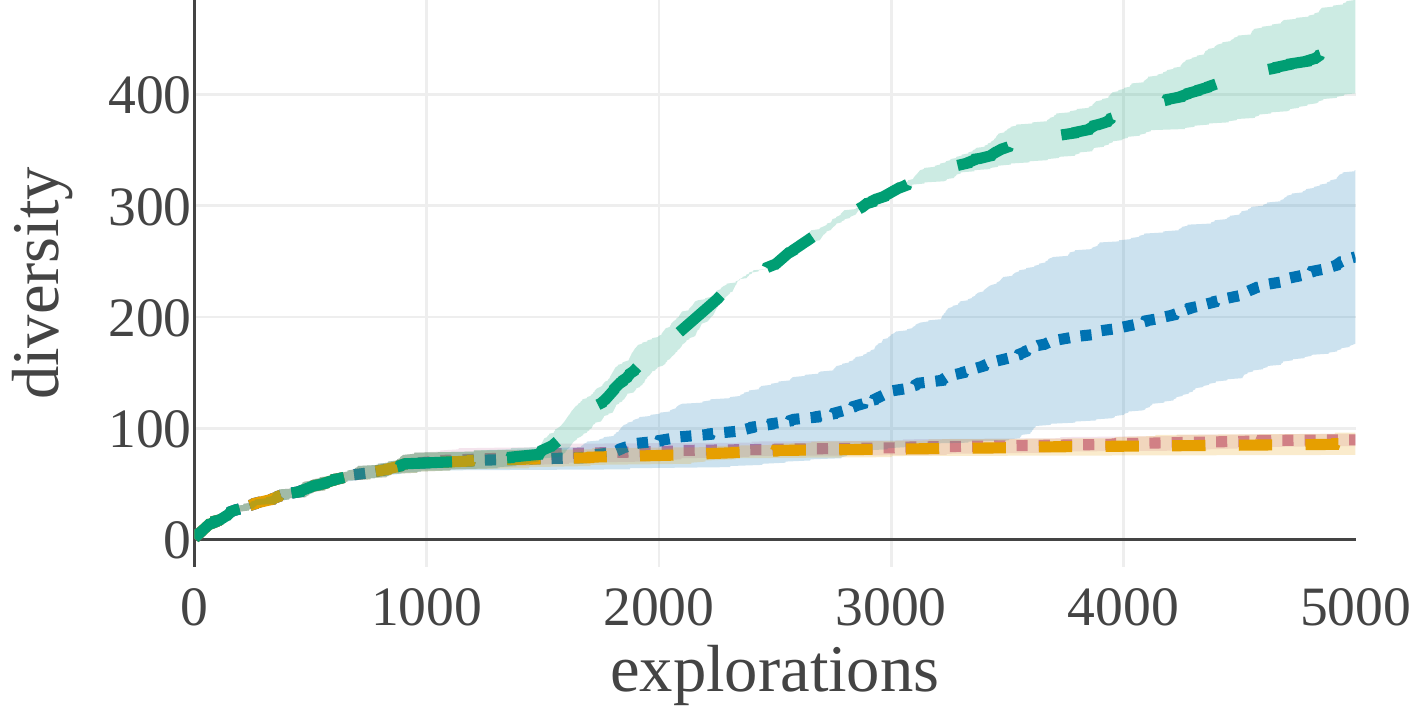} \\
\end{tabular}
}
 \caption{
 Depicted is the average diversity ($n=5$) with the standard deviation as shaded area for the different IMGEP variants. The diversity in computed in the goal space learned by the IMGEP-VAE algorithm and in the different goal spaces learned by the IMGEP-HOLMES variant.}
\label{sm:fig_10}
\end{figure}

Additionally to the percentage of identified patterns presented in table \ref{table_1} of the main paper, we provide in this section an analysis of the diversity discovered by the different IMGEP variants.

During the exploration phase, the different IMGEP variants sample in their goal spaces respectively. Single VAE and HOLMES embedding networks have their own goal spaces and their own representations of the explored images. To evaluate the discovered diversity by each variant, we project each set of exploration images to each goal space and calculate the diversity measure defined by \cite{reinke2019intrinsically} in section B.7.1.  The diversity measure is defined as the area covered by the representations of a certain image set in the respective goal space. Additionally,  the goal spaces are divided in bins to simplify the area calculation and the final diversity measure is calculated as the count of goal space bins which have at least one representation point inside. Each axis of goal space is divided in 4 bins (including the out of range areas).
  
Figure \ref{sm:fig_10} shows the diversity measure in each goal space for each exploration strategy (IMGEP-VAE, IMGEP-HOLMES(A), IMGEP-HOLMES(NA)) over time (runs of exploration). Each experiment starts with 1000 steps of random exploration after which goal oriented strategy starts. 

As shown in \ref{sm:fig_9}, VAE goal space and HOLMES 0 root node goal space encode the same type of information, and therefore have similar diversity profiles. For this \enquote{type} of diversity, IMGEP-VAE reaches a higher diversity than HOLMES as the goal-sampling strategy operates in that space during the whole course of exploration and therefore manage to cover it better. However, HOLMES nodes encode different type of information and hence diversity, resulting in different diversity profiles. For a better understanding of the different \enquote{type} of diversity encoded in the different goal spaces, we refer to figure \ref{sm:fig_3} that illustrate the kind of patterns onto which each goal space representation was trained. In HOLMES 000 goal space, every algorithm reaches the same (null) diversity because this space gathers only dead (all-white) patterns which are all encoded to the same feature point. We can see the impact of \textit{guidance} in exploration as IMGEP-HOLMES (A) reaches a higher diversity in the goal spaces 00, 001, 0010 and 0011 of IMGEP-HOLMES and these goal spaces were trained mainly on \enquote{animals} as depicted in figure \ref{sm:fig_3} and figure \ref{sm:fig_4}. Similarly IMGEP-HOLMES (NA) reaches a higher diversity in the goal spaces 011, 0111, 01110 and  01111 of IMGEP-HOLMES and these goal spaces were trained mainly on \enquote{non-animals}.

\section{Implementation Details}
\label{sm:sec_B}
This appendix complements section \ref{sec_2} of the main paper. First we detail the framework of intrinsically-motivated goal exploration processes (IMGEPs). Then we detail the integration of HOLMES into the IMGEP process.

\subsection{Intrinsically Motivated Goal Exploration Processes (IMGEP)}
\label{sm:sec_B_1}
This section retakes the IMGEP formalization of \cite{reinke2019intrinsically}, we refer to this paper for additional details.
An IMGEP is an algorithmic process which automatically generates a sequence of goals in order to explore the parameters of an unknown complex system. It aims to maximize the diversity of observations from that system within a budget of $N$ experiments. IMGEPs are equipped with a memory of past experimental parameters and observations, denoted as the history $\mathcal{H}$, which is used to guide the exploration process. 

The explored system is characterized with three components.
A parameter space $\Theta$ corresponding to the system parameters $\theta$ that are under the agent control. An observation space $O$ where an observation $o$ is a vector representing all the signals captured from the system, in our case raw sensory images of the discovered patterns. The (unknown) environment dynamics $D$: $\Theta \rightarrow O$ mapping parameters to observations.

To explore a system, an IMGEP uses a goal space $\mathcal{G}$ computed by an encoding function $\hat{g} = \mathcal{R}(o)$, where the goal sampling stategy is implemented.

The exploration process iterates through $N$ exploration runs with the following strategy. First, sample a goal $g$ from a goal sampling distribution $G$ defined in $\mathcal{G}$ and based on the history of reached points $\mathcal{H}$. Then, infer corresponding parameter $\theta$ using a parameter sampling policy $\Pi = \Pr(\theta; g, \mathcal{H})$ and roll-out an experiment with $\theta$. Observe the outcome $o$ and compute the corresponding encoding $\mathcal{R}(o)$. Store the experimental parameters, observation and reached goal $(\theta,o ,\mathcal{R}(o))$ in history $\mathcal{H}$.

Because the parameter sampling policy $\Pi$ and the goal sampling distribution $G$ generally take into account previous explorations runs, the history is first populated through exploring $N_{init}$ randomly sampled parameters after which the intrinsically motivated goal exploration process starts.

\subsection{IMGEP-HOLMES}
\label{sm:sec_B_2}
IMGEP-HOLMES replaces the representation $\mathcal{R}$ with the proposed hierarchy of deep generative models $\{\mathcal{R}_k\}$ to encode the observations and goals at different levels along the hierarchy.

Because this new representation creates a hierarchy of modular goal spaces, the goal-sampling strategy is divided in two steps: 1) sample a target goal space $\mathcal{G}_k$ according to a goal space sampling distribution $G_{space}(\mathcal{H})$, 2) sample a target goal $g$ in this space according to a goal sampling distribution $G_k(\mathcal{H})$.

During exploration, when a certain goal space $\mathcal{G}_k$ gets saturated, it is split into two new goal spaces $\mathcal{G}_{k_{left}}$ and $\mathcal{G}_{k_{right}}$. Each goal space inherits from a part of the population of $\mathcal{G}_k$. Two child module representations $\mathcal{R}_{k_{left}}$ and $\mathcal{R}_{k_{right}}$ are instantiated with new neural architecture and randomly initialized.

A pseudo-code for IMGEP-HOLMES implementation is given in algorithm \ref{sm:algo_1}.
\begin{algorithm}[b!]
\SetKwRepeat{Do}{do}{while}
\DontPrintSemicolon
Initialize the goal space representation $\mathcal{R} = \mathcal{R}_0$ with random weights\\
\For{$i\leftarrow 1$ \KwTo $N$}{
\If(\tcp*[f]{Initial random iterations to populate $\mathcal{H}$}){$i<N_{init}$} {
Sample $\theta \sim \mathcal{U}(\Theta)$
}
\Else(\tcp*[f]{Intrinsically motivated iterations}){
Sample a target goal space $\mathcal{G}_k \sim G_{space}(\mathcal{H})$ in the hierarchy \\
Sample a goal $g \sim G_k(\mathcal{H})$ in $\mathcal{G}_k$  \\
Choose $\theta \sim \Pi(\mathcal{G}_k, g, \mathcal{H})$  \\
}
Perform an experiment with $\theta$ and observe $o$ \\
$\mathcal{R}_k \gets \mathcal{R}_0 $\\
\While(\tcp*[f]{Encode reached goals in the hierarchy}){$\mathcal{R}_k$ not a leaf module}{
 $\hat{g} = \mathcal{R}_k(o, \{\mathcal{R}_{ancestors(k)}(o)\})$ \\
 Append $(\theta, o, \hat{g})$ to the history $\mathcal{H}[k]$ \\
  $\mathcal{R}_k \gets child(\mathcal{R}_k | \hat{g}) $\\
 }
\BlankLine
\If(\tcp*[f]{Augment representational capacity}){a goal space $\mathcal{G}_k$ is \textit{saturated}} {
Freeze $\mathcal{R}_k$ weights \\
Define a boundary $B_k$ splitting $\mathcal{G}_k$ in two subspaces $\mathcal{G}_{k_{left}}$ and $\mathcal{G}_{k_{right}}$\\
Instantiate two child modules $\mathcal{R}_{k_{left}}$ and $\mathcal{R}_{k_{right}}$
\For{$(\theta, o, \hat{g}) \in \mathcal{H}[k]$}{ 
\If{$\hat{g}$ is on the left side of $B_k$} {
Append $(\theta, o, \mathcal{R}_{k_{left}}(o))$ to the history $\mathcal{H}[k_{left}]$}
\Else {
Append $(\theta, o, \mathcal{R}_{k_{right}}(o))$ to the history $\mathcal{H}[k_{right}]$
}
}
}
\BlankLine
\If(\tcp*[f]{Periodically train the network}){$i \mod T == 0$} {
\For{E epochs} {Train the hierarchy $\mathcal{R}$ on observations in $\mathcal{H}$ with importance sampling}
\For(\tcp*[f]{Update the database of reached goals}){$k \in \mathcal{H}$}{
\For{$(\theta, o, \hat{g}) \in \mathcal{H}[k]$}{$\mathcal{H}[k][\hat{g}] \gets \mathcal{R}_k(o)$}
}}}
\caption{IMGEP-HOLMES}
\label{sm:algo_1}
\end{algorithm}

In this paper, the following design choices were made to decide when and how to split a node in the hierarchy. When the population of a goal space go past a threshold $N_{max}$, we trigger a split in that space. Other approaches could be considered as trigger signal such as a drop in the reconstruction loss \citep{caselles2019s}, a low increase of diversity progress, etc. We use the reconstruction performance to separate the population in the selected goal space in two: the median reconstruction error serves as threshold to classify the population as \enquote{badly} versus \enquote{well} reconstructed
and a Support Vector Machine (SVM) classifier is then fitted generating \textit{boundary} $B_k$ in the goal space. From that boundary, the frozen node redirects incoming data flow to a certain child module.

\section{Experimental settings}
In this section we detail the experimental settings and hyperparameters.

We refer to the appendix of \cite{reinke2019intrinsically} for Lenia settings (section B.1) and sampling mechanisms for Lenia’s initial state via CPPN and dynamic parameters (section B.4). The same hyperparameters were used in this paper.

Table \ref{sm:table_2} reports the VAE neural network architecture for the IMGEP-VAE representation and table \ref{sm:table_3} reports the neural architecture of the core module for the IMGEP-HOLMES variant. We give a lower capacity to the IMGEP-HOLMES core network (38 600 total number of parameters) than to the IMGEP-VAE network (572 000 total number of parameters). However, the total number of parameters of HOLMES is incrementally augmented each time a new module and its corresponding connections are added in the hierarchy. A possible solution to control the final total number of parameters is to fix a maximum number of splits in advance. 

The networks are trained 400 epochs every 400 runs of exploration, and initialized with kaiming uniform initialization. For HOLMES child modules, the first convolutional layers are initialized with the values of the trained parent module. We used the Adam optimizer ($lr=1\mathrm{e}{-3}$, $\beta_1=0.9$, $\beta_2=0.999$, $\epsilon=1\mathrm{e}{-8}$, weight decay=$1\mathrm{e}{-5}$) with a batch size of 128.

\begin{table}[h]
	\centering
    \resizebox*{1.0\textwidth}{!}{%
	\begin{tabular}{ll}
   & \\
   \textbf{Encoder} & \textbf{Decoder} \\
   \cmidrule(lr){1-1} \cmidrule(lr){2-2}
   Input pattern A: $256\times256\times1$ & Input latent vector z: $16\times1$ \\
   Conv layer: 32 kernels $4\times4$, stride $2$, $1$-padding + ReLU & FC layers : 256 + ReLU,  $16\times16\times32$ + ReLU \\
   Conv layer: 32 kernels $4\times4$, stride $2$, $1$-padding + ReLU & TransposeConv layer: 32 kernels $4\times4$, stride $2$, $1$-padding + ReLU \\
   Conv layer: 32 kernels $4\times4$, stride $2$, $1$-padding + ReLU & TransposeConv layer: 32 kernels $4\times4$, stride $2$, $1$-padding + ReLU \\
   Conv layer: 32 kernels $4\times4$, stride $2$, $1$-padding + ReLU & TransposeConv layer: 32 kernels $4\times4$, stride $2$, $1$-padding + ReLU \\
   Conv layer: 32 kernels $4\times4$, stride $2$, $1$-padding + ReLU & TransposeConv layer: 32 kernels $4\times4$, stride $2$, $1$-padding + ReLU \\
   Conv layer: 32 kernels $4\times4$, stride $2$, $1$-padding + ReLU & TransposeConv layer: 32 kernels $4\times4$, stride $2$, $1$-padding + ReLU \\
   FC layers : 256 + ReLU, 256 + ReLU, FC: $2\times16$ & TransposeConv layer: 32 kernels $4\times4$, stride $2$, $1$-padding \\
\end{tabular}}
    \vspace{0.2cm}
	\caption{VAE architecture used in the IMGEP-VAE variant.}
\label{sm:table_2}
\end{table}

\begin{table}[h]
	\centering
    \resizebox*{1.0\textwidth}{!}{%
	\begin{tabular}{ll}
   & \\
   \textbf{Encoder} & \textbf{Decoder} \\
   \cmidrule(lr){1-1} \cmidrule(lr){2-2}
   Input pattern A: $256\times256\times1$ & Input latent vector z: $16\times1$ \\
   Conv layer: 8 kernels $4\times4$, stride $2$, $1$-padding + ReLU & FC layers : 64 + ReLU, $16\times16\times8$ + ReLU \\
   Conv layer: 8 kernels $4\times4$, stride $2$, $1$-padding + ReLU & TransposeConv layer: 8 kernels $4\times4$, stride $2$, $1$-padding + ReLU \\
   Conv layer: 8 kernels $4\times4$, stride $2$, $1$-padding + ReLU & TransposeConv layer: 8 kernels $4\times4$, stride $2$, $1$-padding + ReLU \\
   Conv layer: 8 kernels $4\times4$, stride $2$, $1$-padding + ReLU & TransposeConv layer: 8 kernels $4\times4$, stride $2$, $1$-padding + ReLU \\
   Conv layer: 8 kernels $4\times4$, stride $2$, $1$-padding + ReLU & TransposeConv layer: 8 kernels $4\times4$, stride $2$, $1$-padding + ReLU \\
   Conv layer: 8 kernels $4\times4$, stride $2$, $1$-padding + ReLU & TransposeConv layer: 8 kernels $4\times4$, stride $2$, $1$-padding + ReLU \\
   FC layers : 64 + ReLU, 64 + ReLU, FC: $2\times16$ & TransposeConv layer: 32 kernels $4\times4$, stride $2$, $1$-padding \\
\end{tabular}}
    \vspace{0.2cm}
	\caption{Basis VAE architecture used in the IMGEP-HOLMES variant.}
\label{sm:table_3}
\end{table}

\end{document}